\def\year{2022}\relax
\documentclass[letterpaper]{article} % DO NOT CHANGE THIS
\usepackage{aaai22}  % DO NOT CHANGE THIS
\usepackage{times}  % DO NOT CHANGE THIS
\usepackage{helvet}  % DO NOT CHANGE THIS
\usepackage{courier}  % DO NOT CHANGE THIS
\usepackage[hyphens]{url}  % DO NOT CHANGE THIS
\usepackage{graphicx} % DO NOT CHANGE THIS
\usepackage{natbib}  % DO NOT CHANGE THIS AND DO NOT ADD ANY OPTIONS TO IT
\usepackage{caption} % DO NOT CHANGE THIS AND DO NOT ADD ANY OPTIONS TO IT
\DeclareCaptionStyle{ruled}{labelfont=normalfont,labelsep=colon,strut=off} % DO NOT CHANGE THIS
\usepackage{epsfig}
\usepackage{amssymb}
\usepackage{makecell}
\usepackage{graphicx}
\usepackage{amsmath}
\usepackage{amsthm}
\usepackage{mathrsfs}
\usepackage{amssymb}
\usepackage{booktabs}
\usepackage{algorithm}
\usepackage{algorithmic}
\usepackage{bm} 
\usepackage{overpic}                                                            
\usepackage{subfigure}
\usepackage{color}
\usepackage{threeparttable}
\usepackage{newfloat}
\usepackage{listings}
\floatstyle{ruled}
\newfloat{listing}{tb}{lst}{}
\floatname{listing}{Listing}

\newtheorem{theorem}{Theorem}
\title{Optimized Potential Initialization for Low-latency Spiking Neural Networks}
\author{
    Tong Bu\textsuperscript{\rm 1, \rm 2},
    Jianhao Ding\textsuperscript{\rm 2},
    Zhaofei Yu\textsuperscript{\rm 1, \rm 2}\thanks{Corresponding author},
    Tiejun Huang\textsuperscript{\rm 1, \rm 2}
}
\affiliations{
    \textsuperscript{\rm 1}Institute for Artificial Intelligence, Peking University\\
    \textsuperscript{\rm 2}Department of Computer Science and Technology, Peking University\\
    putong30@pku.edu.cn, djh01998@stu.pku.edu.cn, \{yuzf12,tjhuang\}@pku.edu.cn
}

\usepackage{bibentry}
\begin{document}

\maketitle

\begin{abstract}
Spiking Neural Networks (SNNs) have been attached great importance due to the distinctive properties of low power consumption,  biological plausibility, and adversarial robustness. The most effective way to train deep SNNs is through ANN-to-SNN conversion, which have yielded the best performance in deep network structure and large-scale datasets. However, there is a trade-off between accuracy and latency. In order to achieve high precision as original ANNs, a long simulation time is needed to match the firing rate of a spiking neuron with the activation value of an analog neuron, which impedes the practical application of SNN. In this paper, we aim to achieve high-performance converted SNNs with extremely low latency (fewer than 32 time-steps). We start by theoretically analyzing ANN-to-SNN conversion and show that scaling the thresholds does play a similar role as weight normalization. Instead of introducing constraints that facilitate ANN-to-SNN conversion at the cost of model capacity, we applied a more direct way by optimizing the initial membrane potential to reduce the conversion loss in each layer. Besides, we demonstrate that optimal initialization of membrane potentials can implement expected error-free ANN-to-SNN conversion. We evaluate our algorithm on the CIFAR-10, CIFAR-100 and ImageNet datasets and achieve state-of-the-art accuracy, using fewer time-steps. For example, we reach top-1 accuracy of 93.38\% on CIFAR-10 with 16 time-steps. Moreover, our method can be applied to other ANN-SNN conversion methodologies and remarkably promote performance when the time-steps is small.
\end{abstract}
%Instead of adding extra constraints and alteration in ANNs
%we show that the initialization of membrane potentials, which are typically chosen to be zero for all neurons, can be optimized to alleviate the trade-off between accuracy and latency.

\section{Introduction}
Spiking Neural Networks (SNNs), as the third generation of Artificial Neural Networks (ANNs)~\cite{maas1997networks}, have attracted great attention in recent years.
Unlike traditional ANNs transmitting information at each propagation cycle, SNNs deliver information through spikes only when the membrane potential reaches the threshold~\cite{gerstner2002spiking}. Due to the event-driven calculation, sparse activation, and multiplication-free characteristics~\cite{roy2019towards},  SNNs have greater energy efficiency than ANNs on neuromorphic chips~\cite{schemmel2010wafer,furber2012overview,merolla2014million,davies2018loihi,pei2019towards}. In addition, SNNs have inherent adversarial robustness. The adversarial accuracy of SNNs under gradient-based attacks is higher than ANNs with the same structure~\cite{sharmin2020inherent}.  Nevertheless, the use of  SNNs is still limited as it remains challenging to train high-performance SNNs.

\begin{figure}[t] 
\includegraphics[width=0.47\textwidth]{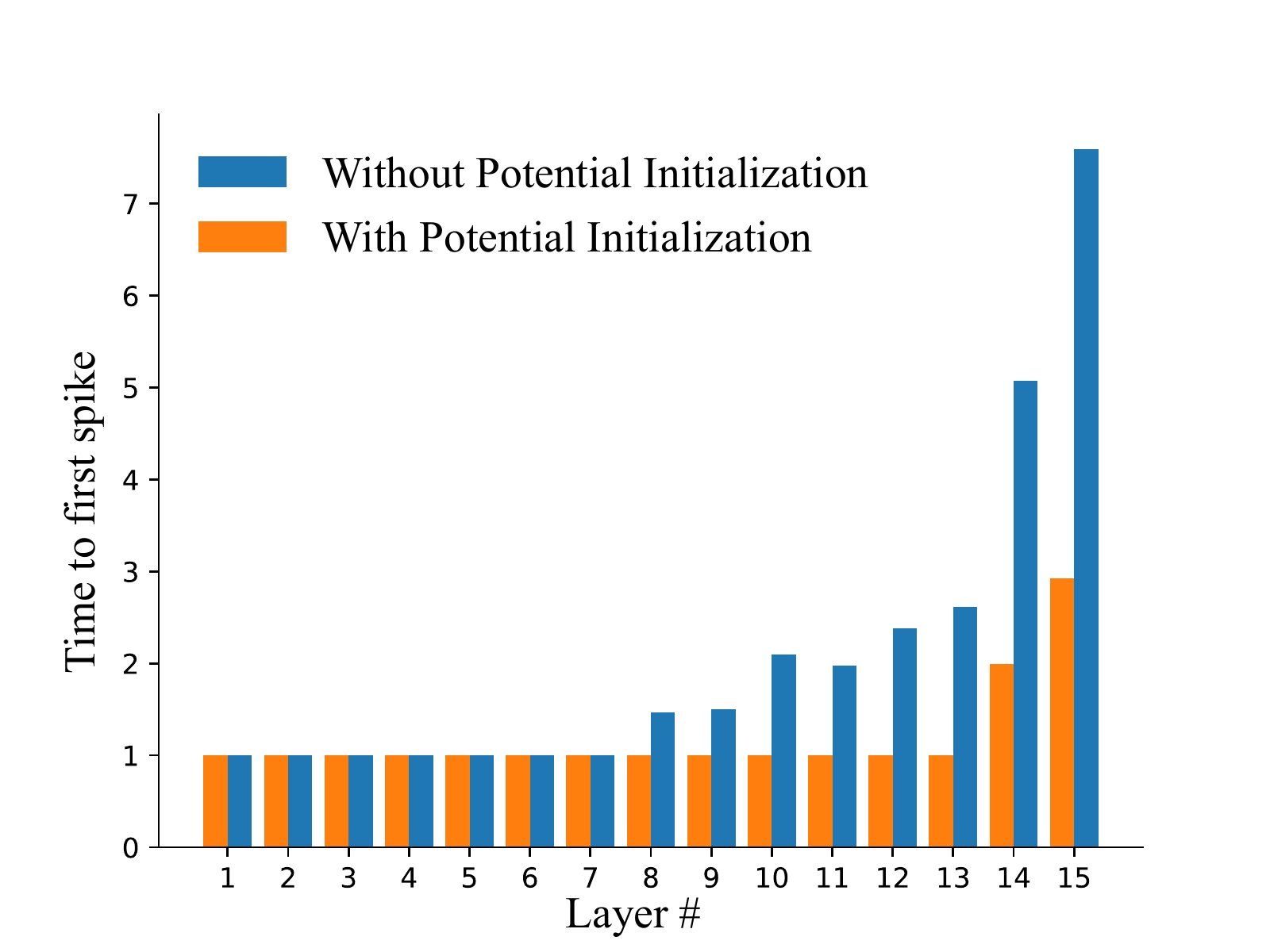}
\caption{Comparison of the propagation delay of converted SNN from VGG-16 with/without membrane potential initialization on the CIFAR-10 dataset. The converted SNN without potential initialization suffers a much longer propagation delay than that with potential initialization.}  
\label{fig1} 
\end{figure}

Generally, there are two main approaches to train a multi-layer SNN:  (1) gradient-based optimization and (2) ANN-to-SNN conversion. The gradient-based optimization takes the idea of ANNs and computes the gradient through backpropagation~\cite{lee2016training,lee2020enabling}.  Although the surrogate gradient methods have been proposed to mitigate the non-differentiable problem of the threshold-triggered firing of SNNs~\cite{shrestha2018slayer, wu2018STBP,neftci2019surrogate}, it is still limited to shallow SNNs as the gradient becomes much unstable when the layer goes deeper~\cite{zheng2021going}. Besides, the gradient-based optimization method requires more GPU computing than ANN training. 

Unlike the gradient-based optimization method, ANN-to-SNN conversion builds the relationship between activation of analog neurons and dynamics of spiking neurons, and then maps the parameters of a well-trained ANN to an SNN with low accuracy loss~\cite{cao2015spiking,diehl2015fast,rueckauer2017conversion,han2020rmp}. Thus high-performance SNNs can be obtained without additional training. ANN-to-SNN conversion requires nearly the same GPU computing and time as ANN training, and has yielded the best performance in deep network structure and large-scale datasets~\cite{deng2020optimal}. Despite these advantages, there has been a trade-off between accuracy and latency. In order to achieve high precision as original ANNs, a long simulation time is needed to match the firing rate of a spiking neuron with the activation value of an analog neuron, which impedes the practical application of SNN.

In this paper, we make a step towards high-performance converted SNNs with extremely low latency (fewer than 32 time-steps).  Instead of introducing constraints that facilitate ANN-to-SNN conversion at the cost of model capacity, we show that the initialization of membrane potentials, which are typically chosen to be zero for all neurons, can be optimized to alleviate the trade-off between accuracy and latency.
%we applied a more direct way by optimizing the initial membrane potential to reduce the conversion loss in each layer. 
% Instead of estimating the layer-wise difference between ANNs and SNNs, and then compensating the error layer-wisely, we show that the initialization of membrane potentials is much more important, which can be optimized to alleviate the trade-off between accuracy and latency.  
Although zero initialization of membrane potentials can make it easier to relate activation of analog neurons to dynamics of spiking neurons,  it also comes with inevitable long latency problems. As illustrated in Fig.~\ref{fig1}, we find that without proper initialization, the neurons in converted SNN take a long time to fire the first spike, and thus the network is ``inactive" in the first few time-steps.
Based on this, we analyze ANN-to-SNN conversion theoretically and prove that the expectation of square conversion error reaches the minimum value when the initial membrane potential is half of the firing threshold. Meanwhile, the expectation of conversion error reaches zero. By setting an optimal initial value in converted SNN, we find a considerable decrease in inference time and a remarkably increased accuracy in low inference time. 

The main contributions of this paper can be summarized as follows:

\begin{itemize}
\item We theoretically analyze ANN-to-SNN conversion and %derive the relationship between the forwarding process of an ANN and the dynamics of an SNN. We 
show that scaling the thresholds does play a similar role as weight normalization, which can help to explain why threshold balancing can reduce the conversion loss and improve the inference latency.
\item %Instead of estimating the layer-wise difference between ANNs and SNNs, and then compensating the error layer-wisely, 
We prove that the initialization of membrane potentials, which are typically chosen to be zero for all neurons, can be optimized to implement expected error-free ANN-to-SNN conversion.
\item We demonstrate the effectiveness of the proposed method in deep network architectures on the CIFAR-10, CIFAR-100 and ImageNet datasets. The proposed method achieves state-of-the-art accuracy on nearly all tested datasets and network structures, using fewer time-steps.
\item We show that our method can be applied to other ANN-SNN conversion methodologies and remarkably promote performance when the time-steps is small. 
\end{itemize}

% In this section, we present a commonly happen phenomenon and explain that phenomenon from a new perspective. In common ANN-SNN conversion methodologies, the converted SNNs often suffers from long inference time to achieve loss-less conversions. We notice that we have to wait a small amount of time before receiving the first spike from the last layer neurons when inferring with converted SNNs, and we always have to wait another period of time until receiving stable spikes from those neurons. 

% Trivially, if we applying threshold balancing on converted SNNs and set the threshold of neurons as maximum value of activation value in ANNs, those neurons need to wait at least one time-step before outputting spikes. Therefore, the propagation delay of this layer is at least one time-step. In a common VGG-16 architecture, as there is 16 layers in total, the least propagation delay is 16 time-steps and the situation may deteriorate in deeper networks.

% One direct cause to that problem is that the all neurons is "inactive" in the first few steps. If we apply proper potential initialization, the neurons can be activated by the initial value and output the first spikes earlier. Therefore, we come up with the method of membrane potential initialization. By setting a initial value between zero and threshold of the neuron in converted SNN, we find a considerable decrease on inference time and a remarkably increase of accuracy in low inference time.

\section{Related Work}

\paragraph{Gradient-based optimization} 
The gradient-based optimization methods directly compute the gradient through backpropagation, which can be divided into two different categories~\cite{kim2020unifying}: (1) activation-based methods and (2) timing-based methods. The activation-based methods unfold the SNNs into discrete time-steps and compute the gradient with backpropagation through time (BPTT), which borrow the idea from training recurrent neural networks in ANNs~\cite{lee2016training,lee2020enabling}. As the gradient of the activation with respect to the membrane potential is non-differentiable, the surrogate gradient is often used~\cite{shrestha2018slayer, wu2018STBP,neftci2019surrogate,ijcai2021-236, fang2021deep,fang2020incorporating}. However, there is a lack of rigorous theoretical analysis of the surrogate gradient~\cite{zenke2021remarkable,zenke2021visualizing}. When the layer of SNNs becomes deeper ($>$50 layers), the gradient becomes much unstable, and the networks suffer the degradation problem~\cite{zheng2021going}. The timing-based methods utilize some approximation methods to estimate the gradient of timings of firing spikes with respect to the membrane potential at the spike timing, which can significantly improves runtime efficiency of BP training. However, they
are usually limited to shallow networks ($<$10 layers)~\cite{mostafa2017supervised,kheradpisheh2020temporal,zhang2020temporal,zhou2021temporal,wu2021tandem}.

\noindent
\textbf{ANN-to-SNN conversion} The ANN-to-SNN conversion is first proposed by Cao et al.~\shortcite{cao2015spiking}, which trains an ANN with ReLU activations and then converts the ANN to an SNN by replacing the activations with spiking neurons. By properly mapping the parameters in ANN to SNN, deep SNNs can gain comparable performance as deep ANNs. Further methods have been proposed to analyze conversion loss and improve the overall performance of converted SNNs, such as weight normalization and threshold balancing~\cite{diehl2015fast, rueckauer2016theory, sengupta2019going}. A soft reset mechanism is applied to IF neurons in previous work~\cite{rueckauer2016theory, han2020rmp}, to avoid information loss when neurons are reset. These works can achieve loss-less conversion with long inference time-steps~\cite{kim2020spiking}, but still suffer from severe accuracy loss with relatively small time-steps. In recent works, most studies focus on accelerating the inference with converted SNN. Stockl and Maass~\shortcite{stockl2021optimized} propose new spiking neurons to better relate ANNs to SNN. Han and Roy~\shortcite{han2020deep} use a time-based encoding scheme to speed up inference. RMP~\cite{han2020rmp}, RNL~\cite{ding2021optimal} and TCL~\cite{ho2020tcl} try to alleviate the trade-off between accuracy and latency by adjusting the threshold dynamically.  Ding et al.~\shortcite{ding2021optimal} propose an optimal fit curve to quantify the fit between ANNs' activations and SNNs' firing rates and demonstrate that the inference time can be reduced by optimizing the upper bound of the fit curve.  Hwang et al.~\shortcite{hwang2021low} proposed a layer-wisely searching algorithm and performed adequate experiments to explore the best initial value of membrane potential. Deng et al.~\shortcite{deng2020rethinking} and Li et al.~\shortcite{li2021free} propose a new method to shift weight, bias and membrane potential in each layer, making relatively low-latency in converted SNNs. Different from the above methods, we directly optimize the initial membrane potential to increase performance at low inference time.

% The ANN-SNN conversion methodology was first raised by \cite{cao2015spiking}. By properly mapping the parameters in ANN to SNN, deep spiking networks can gain comparable performance with deep ANNs. Further methods have been proposed to promote the overall performance of converted SNNs, such as weight normalization and threshold balancing. \cite{diehl2015fast, rueckauer2016theory, sengupta2019going} In work \cite{rueckauer2016theory, han2020rmp}, a soft reset mechanism was applied in IF neurons, avoiding information loss when neuron resets. Those work can achieve loss less conversion with long inference times, but still suffer from serve accuracy loss with inference time is relatively small. In recent works, most studies have focused on accelerating the inference with converted SNN. RMP \cite{han2020rmp} and RNL \cite{ding2021optimal} dynamically adjust the threshold while TSC \cite{han2020deep} use time based encoding scheme to achieve nearly loss-less conversion with small inference time. \cite{deng2020rethinking} and \cite{li2021free} shifting weight and bias in each layer making low-latency converted SNNs. We applied a more direct way by addressing the idea of membrane potential initialization to reduce the conversion loss in each layer. This method can be applied on most ANN-SNN conversion methodologies and remarkably promote performance when inference time is small.

%\clearpage
%\newpage
\section{Methods}
In this section, we first introduce the neuron models for ANNs and SNNs, then we derive the mathematical framework for ANN-to-SNN conversion. Based on this, we show that the initial membrane potential is essential to ANN-to-SNN conversion, and derive the optimal initialization to achieve expected error-free conversion.
\subsection{ANNs and SNNs}
The fundamental idea behind ANN-to-SNN conversion is to build the relationship between the activation value of an analog neuron and the firing rate of a spiking neuron. Based on this relation, we can map the weights of trained ANNs to SNNs. Thus high-performance SNNs can be obtained without additional training~\cite{cao2015spiking}. To be specific, for an ANN, the ReLU activation function of analog neurons in layer $l$ ($l=1,2,...,L$) can be described as:
\begin{align}
    \label{ann}
	\bm{a}^{l} = \max(W^{l}\bm{a}^{l-1}+\bm{b}^{l},0),
\end{align}
where vector $\bm{a}^l$ denotes the output activation values of all neurons in layer $l$, $W^{l}$ is the weight matrix between neurons in layer $l-1$ and neurons in layer $l$, and $\bm{b}^{l}$ refers to the bias of the neurons in layer $l$.

For SNNs, we consider the Integrate-and-Fire (IF) model, which is commonly used in the previous works \cite{cao2015spiking,diehl2015fast,han2020rmp}. In the IF model, if the spiking neurons in layer $l$ receive input $\bm{x}^{l-1}(t)$ at time $t$, the temporal membrane potential $\bm{v}_{temp}^{l}(t)$ can be formulated as the addition of its membrane potential $\bm{v}^{l}(t-1)$ at time $t-1$ and the summation of weighted input:
\begin{align}
    \label{snn}
    	\bm{v}_{temp}^{l}(t) =\bm{v}^{l}(t-1)+ W^{l}\bm{x}^{l-1}(t)+\bm{b}^{l},
\end{align}
where $\bm{x}^{l-1}(t)$ denotes the unweighted postsynaptic
potentials from presynaptic neurons in layer $l-1$ at time $t$, $W^{l}$ is the synaptic weights,
%between spiking neurons in layer $l-1$ and neurons in layer $l$
and $\bm{b}^{l}$ is the bias potential of spiking neurons in layer $l$. When any element $\bm{v}_ {temp,i}^{l}(t)$ of $\bm{v}_{temp}^{l}(t)$ exceeds the firing threshold $V^l_{th}$ at layer $l$, the neuron will elicit a spike with unweighted postsynaptic potential $V^l_{th}$:
\begin{align}
    \label{fire}
\bm{s}_i^l\left( t \right) &=\begin{cases}
	1, ~~ \mathrm{if}~~~\bm{v}_{temp,i}^{l}(t)>V_{th}^{l}\\
	0,  ~~\mathrm{otherwise}\\
\end{cases},\\
\bm{x}^{l}(t)&=\bm{s}^l\left( t \right)V^l_{th}. \label{fire2}
\end{align}
Here $\bm{s}_i^l(t)$ is the $i$-th element of $\bm{s}^l(t)$, which denotes the output spike at time $t$ and equals 1 if there is a spike and 0 otherwise. 
After firing a spike, the membrane potential $\bm{v}^{l}(t)$ at the next time-step $t$ will go back to a reset value. Two approaches are commonly used to reset the potential: ``reset-to-zero'' and ``reset-by-subtraction''. As there exists obvious information loss in ``reset-to-zero''\cite{rueckauer2017conversion,han2020deep}, we adopt ``reset-by-subtraction'' mechanism in this paper. Specifically, after the firing, the membrane potential is reduced by an amount equal to the firing threshold  $V^l_{th}$. Thus the membrane potential updates according to:
\begin{align}
    \label{snnnew}
    	\bm{v}^{l}(t) =\bm{v}^{l}(t-1)+ W^{l}\bm{x}^{l-1}(t)+\bm{b}^{l}-\bm{s}^l\left( t \right)V^l_{th}.
\end{align}

\subsection{Theory for ANN-SNN conversion}
In order to relate the firing rate of SNNs to the activation value of ANNs, here we accumulate Eq.~\eqref{snnnew} from time $1$ to $T$, divide it by $T V^l_{th}$, and get:
\begin{align}
    \label{conversation1}
    \frac{\bm{v}^{l}(T)}{T V^l_{th}}-\frac{\bm{v}^{l}(0)}{T V^l_{th}}=\frac{ W^{l} \sum_{t=1}^{T}\bm{x}^{l-1}(t)}{T V^l_{th}}+\frac{\bm{b}^{l}}{V^l_{th}}-\frac{ \sum_{t=1}^{T}\bm{s}^l\left( t \right)}{T}.
\end{align}
We use $\bm{r}^{l}(T)=\frac{ \sum_{t=1}^{T}\bm{s}^l\left( t \right)}{T}$ to denote the firing rates of spiking neurons in layer $l$ during the period from time $0$ to $T$, and substitute Eq.~\eqref{fire2} into Eq.~\eqref{conversation1} to eliminate $\bm{x}^{l-1}(t)$, we have:
\begin{align}
    \label{conversation2}
    	\bm{r}^{l}(T) =\frac{ W^{l}\bm{r}^{l-1}(T) V^{l-1}_{th}+ \bm{b}^{l}}{V^l_{th} }-\frac{\bm{v}^{l}(T)}{T V^l_{th}}+\frac{\bm{v}^{l}(0)}{T V^l_{th}}.
\end{align}

Note that Eq.~\eqref{conversation2} is the core equation of ANN-SNN conversion. 
%In fact, when $T$ grows larger, terms $-\frac{\bm{v}^{l}(T)}{T V^l_{th}}+\frac{\bm{v}^{l}(0)}{T V^l_{th}}$ can be neglected. 
It describes the relationship of the firing rates of neurons in adjacent layers of an SNN, and can be related to the forwarding process of an ANN (Eq.~\eqref{ann}). To see this, we make the following assumption: % The initial membrane membrane potential $\bm{v}^{l}(0)=0$; (b) 
The inference time (latency) $T$ is large enough so that $\frac{\bm{v}^{l}(T)}{T}\approx 0$ and $\frac{\bm{v}^{l}(0)}{T}\approx 0$. Hence Eq.~\eqref{conversation2} can be simplified as:
\begin{align}
    \label{conversation3}
    	\bm{r}^{l}(T) &=\frac{ W^{l}\bm{r}^{l-1}(T) V^{l-1}_{th}+ \bm{b}^{l}}{V^l_{th} } \\&=\max \left (\frac{ W^{l}\bm{r}^{l-1}(T) V^{l-1}_{th}+ \bm{b}^{l}}{V^l_{th} }, 0 \right ). \nonumber
\end{align}
The last equality holds as the firing rate $\bm{r}^{l}(T)$ is strictly restricted in $[0,1]$.
By contrast, the ReLU activation values $\bm{a}^{l}$ of ANNs in Eq.~\eqref{ann} only need to satisfy $\bm{a}^{l}\geqslant 0 $. In fact, the activation values $\bm{a}^{l}$ of ANNs have an upper bound for countable limited dataset. Thus we can perform normalization for all activation values in Eq.~\eqref{ann}. Specifically, assuming that $\bm{z}^{l}=\frac{\bm{a}^{l}}{\max \{\bm{a}^{l} \}}$, where $\max \{\bm{a}^{l} \}$ denotes the maximum value of $\bm{a}^{l}$, we can rewritten Eq.~\eqref{ann} as:
\begin{align}
    \label{annnew}
	\bm{z}^{l} = \max \left (\frac{W^{l}\bm{z}^{l-1} \max \{\bm{a}^{l-1}\} +\bm{b}^{l}}{\max \{\bm{a}^{l} \}},0 \right).
\end{align}

By comparing Eq.~\eqref{conversation3} and Eq.~\eqref{annnew},  we can find an ANN can be converted to an SNN by copying both weights and biases, and setting the firing threshold $V^{l}_{th}$ equal to the upper bound $\max \{\bm{a}^{l} \}$ of the ReLU activation of analog neurons.
Our result can help to explain the previous finding that scaling the firing thresholds can reduce the conversion loss and improve the inference latency~\cite{han2020deep}.  Actually, scaling the thresholds does play a similar role as the weight normalization technique~\cite{diehl2015fast,rueckauer2017conversion} used in ANN-to-SNN conversion.

From another perspective, we can directly relate the postsynaptic potentials of spiking neurons in adjacent layers to the forwarding process of an ANN (Eq.~\eqref{ann}). If we use $\hat{\bm{a}}^{l}(T)=\bm{r}^{l}(T) V^l_{th}= \frac{ \sum_{t=1}^{T} V^l_{th}  \bm{s}^l\left( t \right)}{T}=\frac{ \sum_{t=1}^{T} \bm{x}^l\left( t \right) }{T} $ to denote the average  postsynaptic potentials from presynaptic neurons in layer $l-1$ during the period from time $0$ to $T$ and substitute it to Eq.~\eqref{conversation2},
we have:
\begin{align}
    \label{potential-1}
    	\hat{\bm{a}}^{l}(T) &=W^{l}\hat{\bm{a}}^{l-1}(T)+ \bm{b}^{l}-\frac{\bm{v}^{l}(T)}{T }+\frac{\bm{v}^{l}(0)}{T }\\
    	&= \max \left (W^{l}\hat{\bm{a}}^{l-1}(T)+ \bm{b}^{l}-\frac{\bm{v}^{l}(T)}{T }+\frac{\bm{v}^{l}(0)}{T },0 \right). \nonumber
\end{align}
By comparing Eq.~\eqref{ann} and Eq.~\eqref{potential-1}, we get the same conclusion that if 
%the initial membrane potential $\bm{v}^{l}(0)$ is zero and  
the inference time (latency) $T$ is large enough, an ANN can be converted to an SNN by coping both the weights and the biases. Note that although $V_{th}^l$ is not included in Eq.~\eqref{potential-1}, it also should equal the maximum value of $\bm{a}^{l}$ due to $\hat{\bm{a}}^{l}(T)=\bm{r}^{l}(T) V^l_{th}$.

\subsection{Optimal initialization of membrane potentials}
The exact equivalence between the forwarding process of an ANN and the firing rates (or postsynaptic potentials) of adjacent layers of an SNN discussed above depends on the assumed condition that the time $T$ is large enough so that $\frac{\bm{v}^{l}(T)}{T}\approx 0$ and $\frac{\bm{v}^{l}(0)}{T}\approx 0$. 
It incurs a long simulation time for SNNs when applied to complicated datasets. Moreover, under low latency constraints, there exists an intrinsic difference between $\bm{r}^{l}(T)$ and $\bm{z}^{l}$, which will transfer layer by layer, resulting in considerable accuracy degradation for converted SNNs. In this subsection, we will
analyze the impact of membrane potential initialization and show that optimal initialization can implement expected error-free ANN-to-SNN conversion.

According to Eq.~\eqref{potential-1}, 
%if we use $\hat{\bm{a}}^{l}(T)=\bm{r}^{l}(T) V^l_{th}= \frac{ \sum_{t=1}^{T} V^l_{th}  \bm{s}^l\left( t \right)}{T}=\frac{ \sum_{t=1}^{T} \bm{x}^l\left( t \right) }{T} $ to denote the average  postsynaptic potentials from presynaptic neurons in layer $l-1$ during the period from time $0$ to $T$,
we can rewrite the relationship of the postsynaptic potentials (or firing rates) of spiking neurons in adjacent layers in a new way:
% \begin{align}
%     \label{conversationnew}
%     	\hat{\bm{a}}^{l}(T) =W^{l}\hat{\bm{a}}^{l-1}(T)+ \bm{b}^{l}-\frac{\bm{v}^{l}(T)}{T }+\frac{\bm{v}^{l}(0)}{T }.
% \end{align}
\begin{align}
    \label{conversationpotnew}
    \hat{\bm{a}}^{l}(T) =\max \left ( \frac{V^l_{th}}{T} \left \lfloor \frac{ T W^{l}\hat{\bm{a}}^{l-1}(T)+ T \bm{b}^{l}+\bm{v}^{l}(0)}{ V^l_{th} } \right \rfloor, 0 \right ).
\end{align}
\begin{align}
    \label{conversationnew}
    	{\bm{r}}^{l}(T) = \max \left ( \frac{1}{T} \left \lfloor \frac{ T W^{l}\bm{r}^{l-1}(T) V^{l-1}_{th}+ T \bm{b}^{l}+\bm{v}^{l}(0)}{ V^l_{th} } \right \rfloor, 0 \right ).
\end{align}
%The detailed derivation are presented in \textbf{Supplementary Material B}. 
Here $T W^{l}\hat{\bm{a}}^{l-1}(T)+ T \bm{b}^{l}+\bm{v}^{l}(0)$ represents the accumulated potential from time 0 to $T$. $\lfloor  x \rfloor$ denotes the floor function, which calculates the maximum integer that is smaller or equal to $x$.
Eq.~\eqref{conversationnew} holds as $\hat{\bm{a}}^{l}(T)=\bm{r}^{l}(T) V^l_{th}$.
By comparing Eq.~\eqref{conversationpotnew} and Eq.~\eqref{ann}, Eq.~\eqref{conversationnew} and Eq.~\eqref{annnew}, we can find there exist inherent quantization errors between ANNs and SNNs due to the discrete characteristic of the firing rate. Here we propose a simple and direct way that optimize the initialization of membrane potentials to reduce the error and improve the conversion.
%Instead of estimating the layer-wise difference between ANNs and SNNs, and then compensating the error layer-wisely, we directly optimize the initialization of membrane potentials to reduce the error.
% find that the initialization of membrane potentials % which are typically chosen to be zero for all neurons,  
% can be optimized to implement expected error-free ANN-to-SNN conversions. 

To be specific, we suppose that the ReLU activation $\bm{a}^{l-1}$ in layer $l-1$ of an ANN is the same as the postsynaptic potentials $\hat{\bm{a}}^{l-1}(T)$ in layer $l-1$ of an SNN, that is $\bm{a}^{l-1}=\hat{\bm{a}}^{l-1}(T)$, $W^{l}\bm{a}^{l-1}+\bm{b}^{l}=W^{l}\hat{\bm{a}}^{l-1}(T)+ \bm{b}^{l}$, and then compare the outputs of ANN and SNN in layer $l$. For the convenience of representation, we use $f(\bm{z})= \max(\bm{z},0)$ to denote the activation function
$\bm{a}^{l} =\max(W^{l}\bm{a}^{l-1}+\bm{b}^{l},0)$ of ANNs, that is $\bm{a}^{l}=f(\bm{z})= \max(\bm{z},0)$ and $\bm{z}=W^{l}\bm{a}^{l-1}+\bm{b}^{l}$. Besides, we use $f^{'} \left ({\bm{z}} \right )=\max \left ( \frac{V^l_{th}}{T} \left \lfloor \frac{ T \bm{z} +\bm{v}^{l}(0)}{ V^l_{th} } \right \rfloor, 0 \right )$
to denote the activation function $\hat{\bm{a}}^{l}(T) =\max \left ( \frac{V^l_{th}}{T} \left \lfloor \frac{ T W^{l}\hat{\bm{a}}^{l-1}(T)+ T \bm{b}^{l}+\bm{v}^{l}(0)}{ V^l_{th} } \right \rfloor, 0 \right )$ of SNN, that is 
$\hat{\bm{a}}^{l}(T) =f^{'} \left ({\bm{z}} \right )=\max \left ( \frac{V^l_{th}}{T} \left \lfloor \frac{ T \bm{z} +\bm{v}^{l}(0)}{ V^l_{th} } \right \rfloor, 0 \right )$ and $\bm{z}=W^{l}\hat{\bm{a}}^{l-1}(T)+ \bm{b}^{l}$.
% $\bm{a}^{l} =f(\bm{z})= \max(\bm{z},0)=\max(W^{l}\bm{a}^{l-1}+\bm{b}^{l},0)$ to denote the activation function of ANNs, and $\hat{\bm{a}}^{l}(T) =f^{'} \left ({\bm{z}} \right )=\max \left ( \frac{V^l_{th}}{T} \left \lfloor \frac{ T \bm{z} +\bm{v}^{l}(0)}{ V^l_{th} } \right \rfloor, 0 \right )=\max \left ( \frac{V^l_{th}}{T} \left \lfloor \frac{ T W^{l}\hat{\bm{a}}^{l-1}(T)+ T \bm{b}^{l}+\bm{v}^{l}(0)}{ V^l_{th} } \right \rfloor, 0 \right )$ to denote the activation function of SNN, where $\bm{z}=W^{l}\hat{\bm{a}}^{l-1}(T)+ \bm{b}^{l}=W^{l}\bm{a}^{l-1}+\bm{b}^{l}$. 

The expected squared difference between $\bm{a}^{l}$ and $\hat{\bm{a}}^{l}(T)$ can be defined as:
\begin{algorithm}[t]
	\caption{Overall algorithm of ANN to SNN conversion.}
	\label{algo:training}
	\textbf{Input}:An ANN $f_{\text{ANN}}(\bm{x};{W},\bm{b})$; A Dataset $D$\\
	\textbf{Parameter}: ANN parameters ${W}, \bm{b}$, Trainable clipping upper-bound $\bm{\theta}$\\
    \textbf{Output}: $f_{\text{SNN}}$
	\begin{algorithmic}[1]
	    \FOR{$l = 1$ to $L$}
	    \STATE Replace $\text{ReLU}(\bm{x})$ by $\text{clip}(\bm{x}; 0, \theta^l)$
	    \IF {is MaxPooling}
	    \STATE Replace MaxPooling by AvgPooling
	    \ENDIF
		\ENDFOR
		\FOR{$e = 1$ to $epochs$}
	    \FOR{length of Dataset $D$}
	    \STATE Sample minibatch $(\bm{x}^0,\bm{y}^0)$ from $D$
	    \FOR{$l = 1$ to $L$}
	    \STATE $\bm{x}^{l} = \text{clip}({W}^l\bm{x}^{l-1}+\bm{b}^l , 0, \theta^l)$
	    \ENDFOR
	    \STATE Loss = $\mathscr{L}(\bm{x}^{L};\bm{y})$
		\STATE Update ${W}, \bm{b}, \bm{\theta}$ via stochastic gradient descent
	    \ENDFOR
		\ENDFOR
		\FOR{$l = 1$ to $L$}
		\STATE $f_{\text{SNN}}.{W}^{l}=f_{\text{ANN}}.{W}^{l}$
	    \STATE $f_{\text{SNN}}.\bm{b}^{l}=f_{\text{ANN}}.\bm{b}^{l}$
	    \STATE $f_{\text{SNN}}.V_{th}^l=f_{\text{ANN}}.\theta^{l}$
	    \STATE $f_{\text{SNN}}.\bm{v}^l(0)=f_{\text{SNN}}.V_{th}^l/2$
	    \ENDFOR
	    \RETURN $f_{\text{SNN}}$
	\end{algorithmic}
\end{algorithm}
\begin{align}
    \label{expectation}
&E_{\bm{z}}\left\| f\left( \bm{z} \right) -f^{'}\left( \bm{z} \right) \right\| _{2}^{2} = E_{\bm{z}}\left\|\bm{z} - \frac{V^l_{th}}{T} \left \lfloor \frac{ T \bm{z} +\bm{v}^{l}(0)}{ V^l_{th} } \right \rfloor \right\| _{2}^{2}. 
%\quad  \bm{z} \in [0, V^l_{th}]  
\end{align}
Note that as the threshold is set to be the max activation value of ANN, $\bm{z}=W^{l}\bm{a}^{l-1}+\bm{b}^{l}=\bm{a}^{l}$ should  always fall into interval 
$[0, V^l_{th}]$.
%the post synapse potential should be the same for both ANN and converted SNN and always fall into interval  $[0, V^l_{th}]$. 
If we assume that ${\bm{z}_i}$ is uniformly distributed in every small interval $[m_{t}, m_{t+1}]$ with the probability density function $p_i^t$ ($t=0,1,...,T-1,T$), where $ \bm{z}_i$ and $\bm{v}^{l}_i(0)$ denote the $i$-th element in $ \bm{z}$ and $\bm{v}^{l}(0)$, respectively, $m_{0}=0, m_{T+1}=V^l_{th}, m_t=\frac{t V^l_{th}-\bm{v}_i^{l}(0)}{T}$ for $t=1,2,...,T$, and $p_i^0=p_i^{T}$, then we can obtain the optimal initialization of membrane potentials. We have the following Theorem.

\begin{theorem}
The expectation of square conversion error (Eq.~\eqref{expectation}) reaches the minimum value when the initial value $\bm{v}^{l}(0)$ is ${ \bm{V}^l_{th}}/2$, meanwhile the expectation of conversion error reaches $0$, that is:
\begin{align}
 % \min_{\bm{v}^{l}(0)} E_{\bm{z}}\left\| f\left( \bm{z} \right) -f^{'}\left( \bm{z} \right) \right\| _{2}^{2}&= \frac{{ V^l_{th}}^2}{12T^2},  \\  \bm{v}^{l \ast}(0) = 
 \arg \min_{\bm{v}^{l}(0)} E_{\bm{z}}\left\| f\left( \bm{z} \right) -f^{'}\left( \bm{z} \right) \right\| _{2}^{2}&= \frac{\bm{V}^l_{th}}{2}, \\
\left.
E_{\bm{z}}\left ( f\left( \bm{z} \right) -f^{'}\left( \bm{z} \right) \right )\right|_{\bm{v}^{l }(0)=\frac{\bm{V}^l_{th}}{2}} &=\bm{0}.
\end{align}
\end{theorem}
Here $\bm{V}^l_{th}$ is the vector of $V^l_{th}$. The detailed proof is in the Appendix section. Theorem 1 implies that when $\bm{v}^{l }(0)=\bm{V}^l_{th}/2$, not only the expectation of square conversion error is minimized, but the expectation of error will be zero as well. Thus optimal initialization of membrane potential can implement expected error-free ANN-to-SNN conversion.

\begin{figure}[t]
 \centering
 \subfigure[VGG-16 on CIFAR-10]{\includegraphics[width=0.22\textwidth]{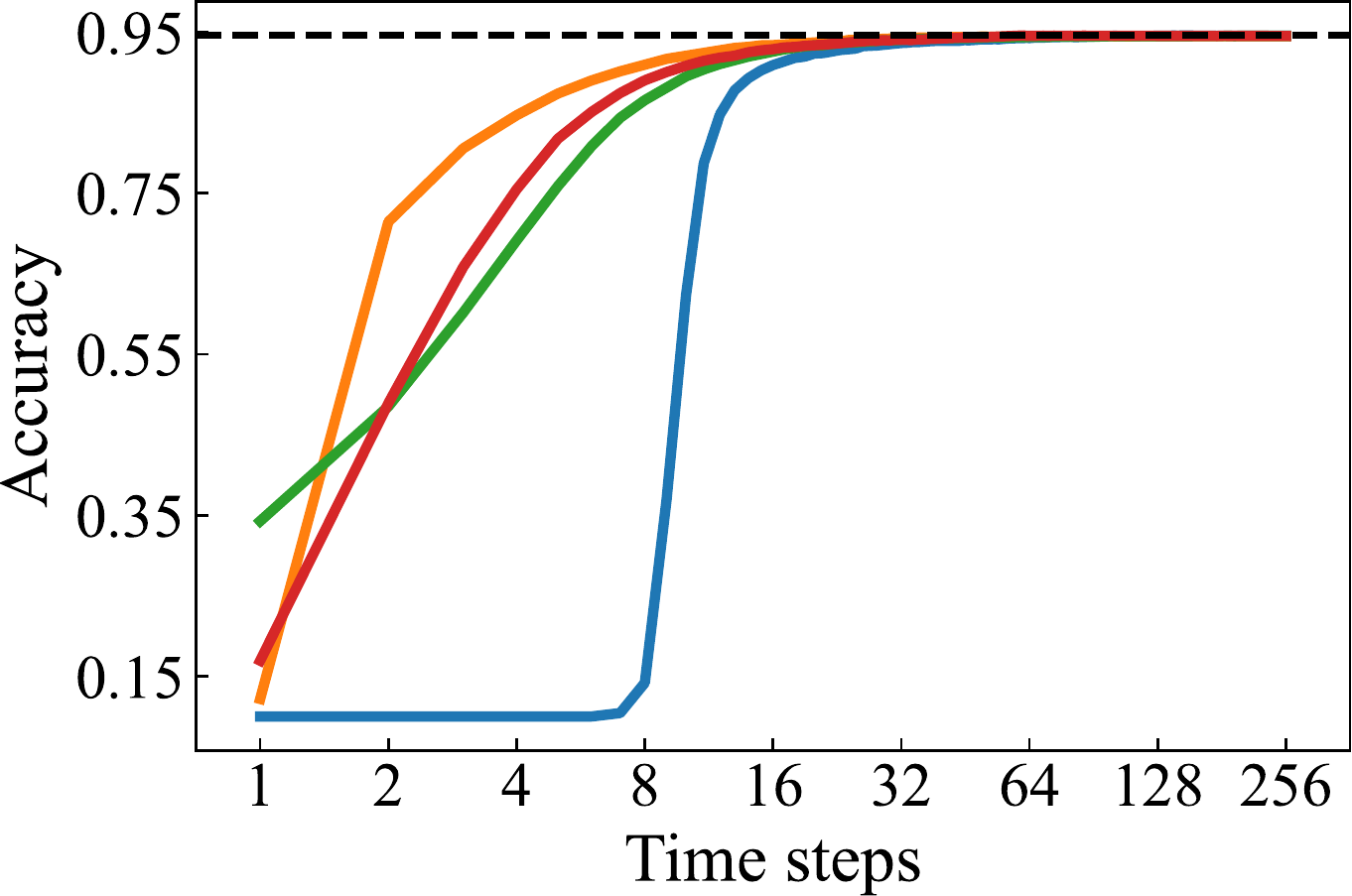}}
 \subfigure[ResNet-20 on CIFAR-10]{\includegraphics[width=0.22\textwidth]{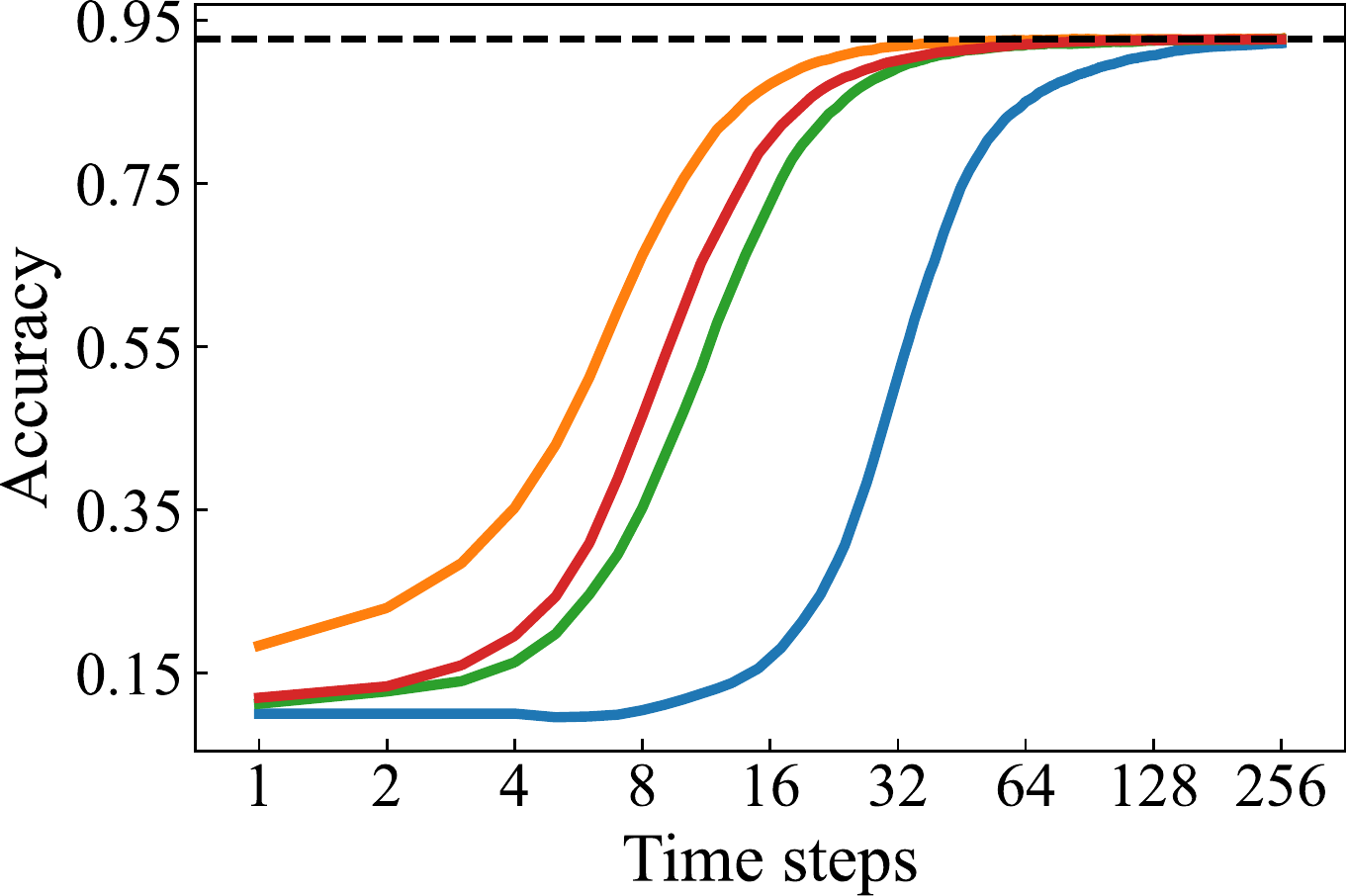}}
 \subfigure{\includegraphics[width=0.4\textwidth]{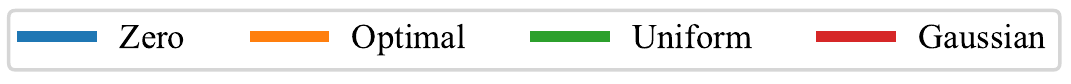}}
 \addtocounter{subfigure}{-1}
 \subfigure[VGG-16 on CIFAR-100]{\includegraphics[width=0.22\textwidth]{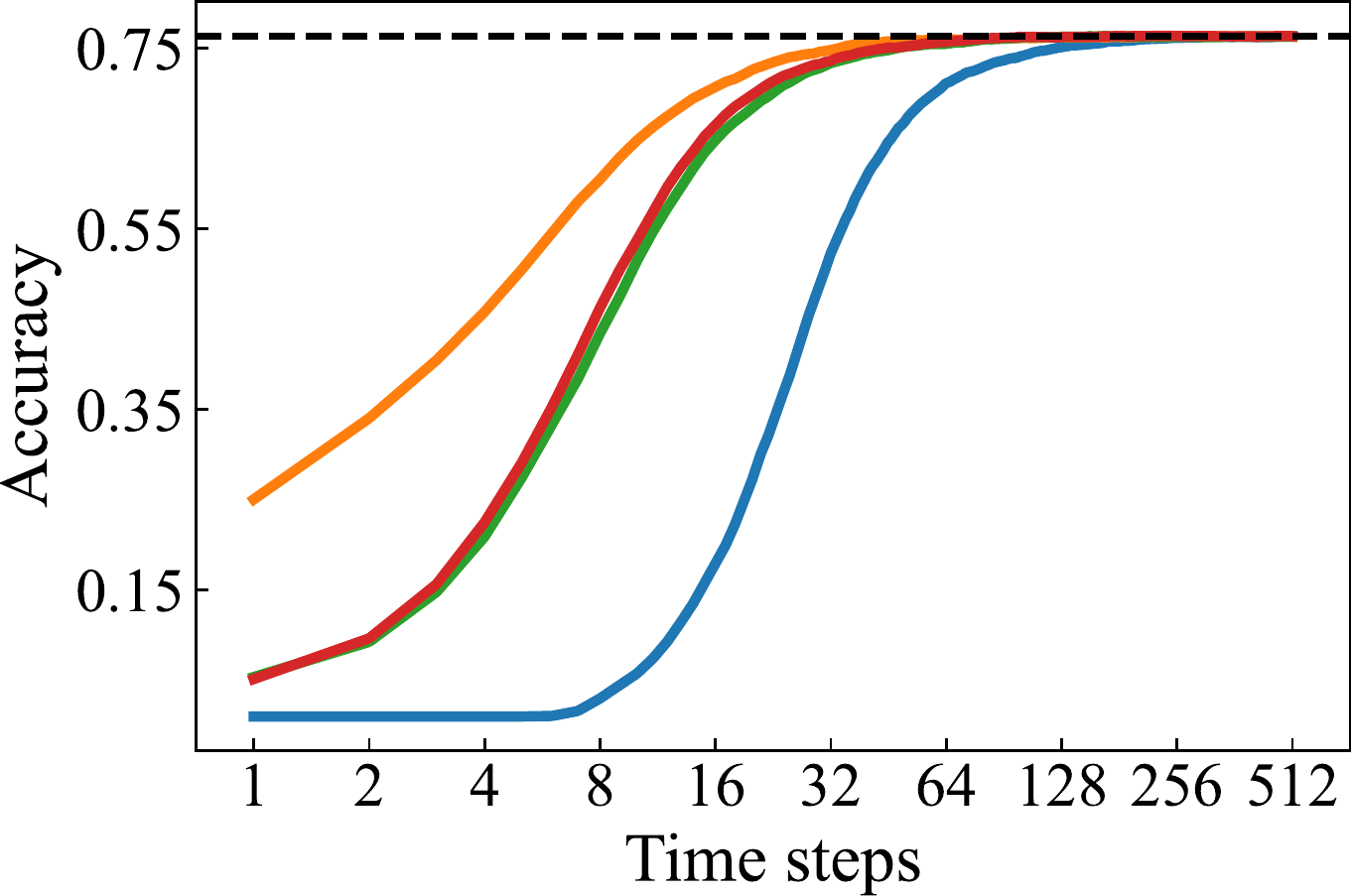}}
 \subfigure[ResNet-20 on CIFAR-100]{\includegraphics[width=0.22\textwidth]{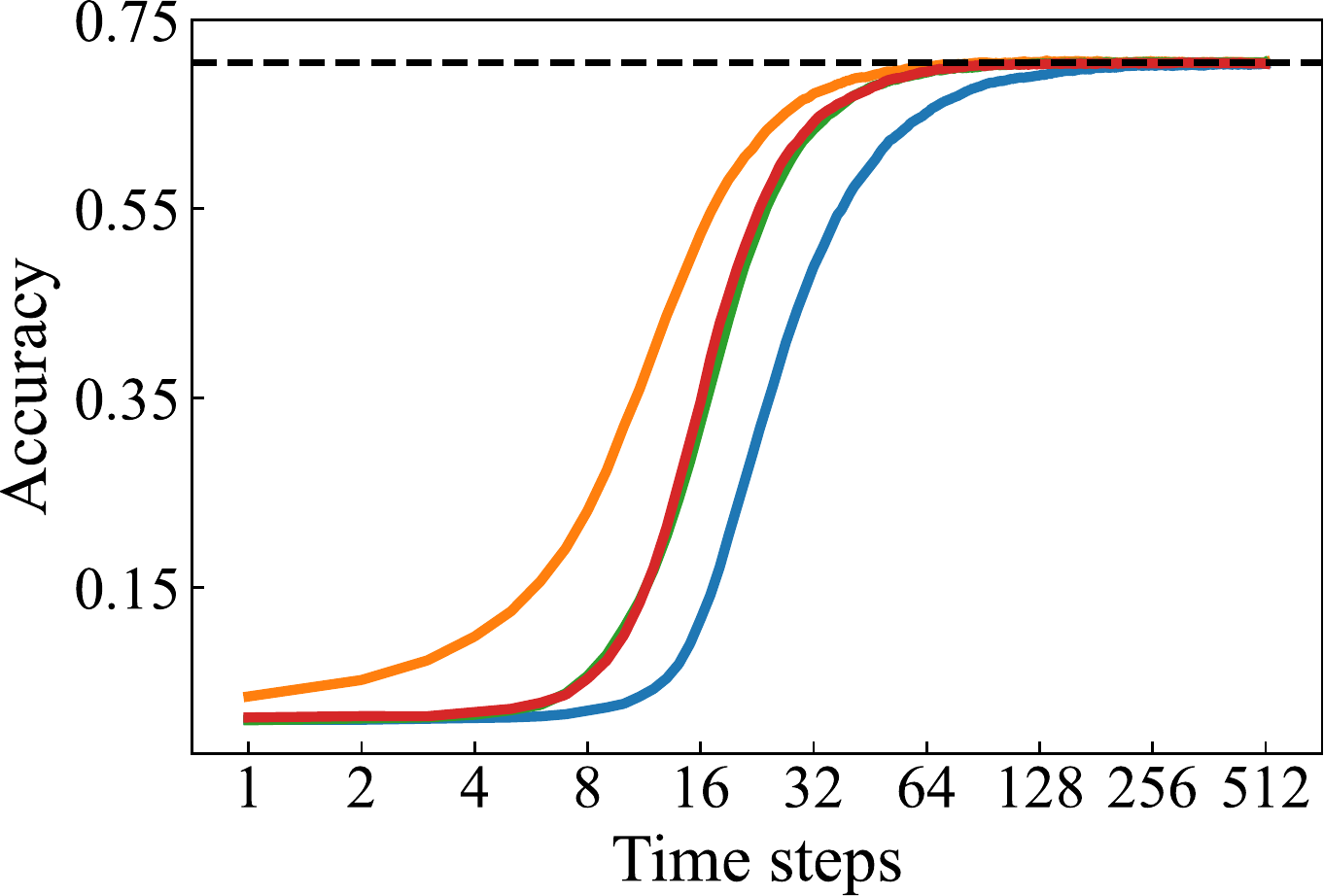}}
 \caption{Comparison of different membrane potential initialization strategies with VGG-16/ResNet-20 network structures on CIFAR-10/CIFAR-100 datasets. The dotted line represents the accuracy of source ANN.}
 \label{Figure 2}
\end{figure}
\section{Experiments}
\subsection{Implementation details}
We evaluate the performance of our methods for classification tasks on CIFAR-10, CIFAR-100 and ImageNet datasets.
For comparison, we utilize VGG-16, ResNet-20 and ResNet-18 network structures as previous work. The proposed ANN-to-SNN conversion algorithm is given in Algorithm~\ref{algo:training}.
For the source ANN, we replace all max-pooling layers with average-pooling layers. In addition, similar to \cite{ho2020tcl}, we add trainable clipping layers to the source ANNs, enabling a better set of the firing thresholds of converted SNNs. For the SNN, we copy both weights and biases from source ANN, and set the firing threshold $V^{l}_{th}$ ($l=1,2,...,L$) equal to the upper bound of the activation of analog neurons. Besides, the initial membrane potential $\bm{v}^{l }(0)$ of all spiking neurons in layer $l$ is set to the same optimal value ${V}^l_{th}/2$. The details of the pre-processing, parameter configuration, and training are as follows.
% More details of the pre-processing, parameter configuration, and training details can be found in Supplementary Material.

\begin{figure}
 \centering
 \subfigure[VGG-16 on CIFAR-10]{\includegraphics[width=0.226\textwidth]{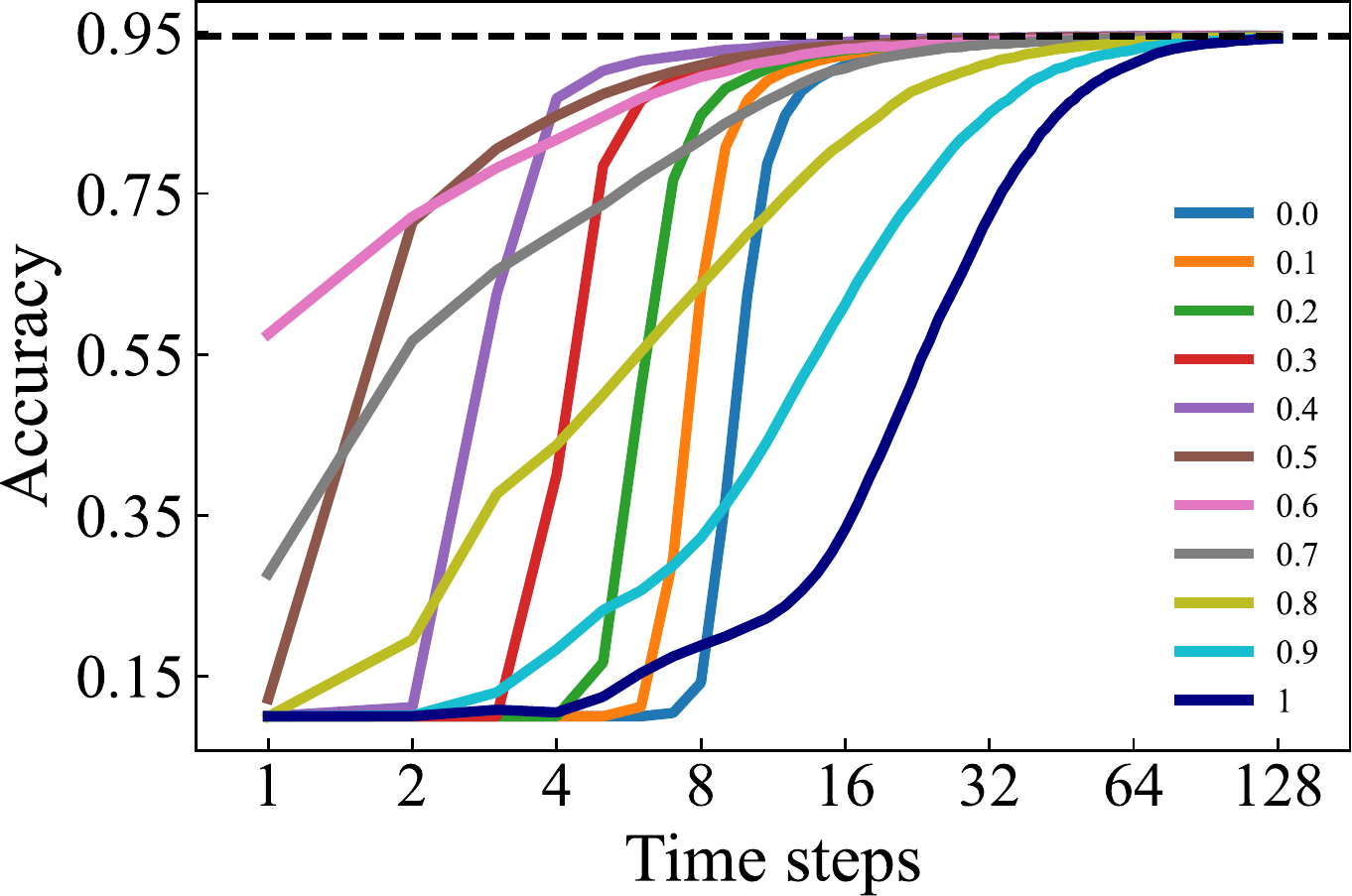}}
 \subfigure[ResNet-20 on CIFAR-10]{\includegraphics[width=0.226\textwidth]{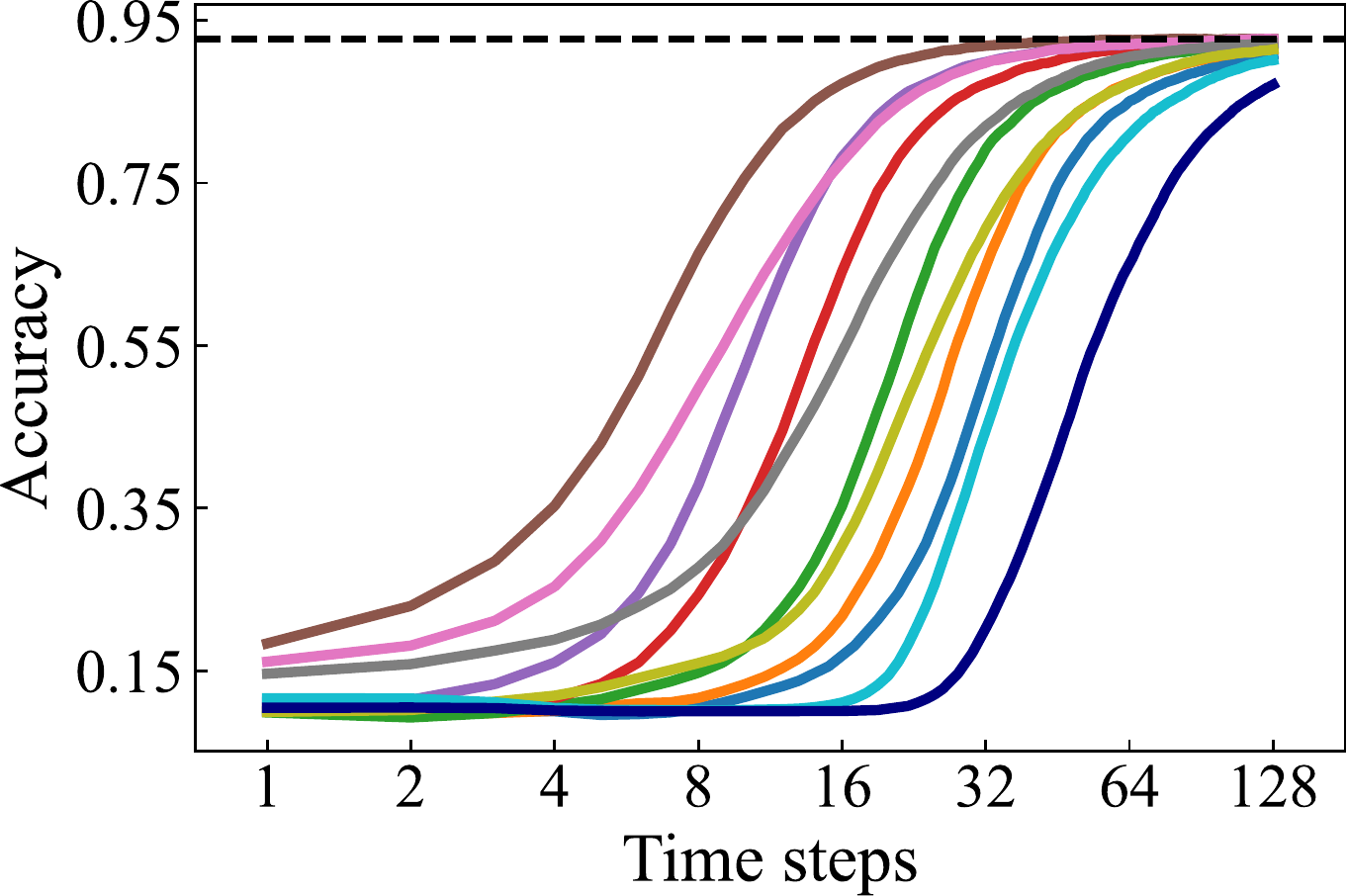}}
 \subfigure[VGG-16 on CIFAR-100]{\includegraphics[width=0.226\textwidth]{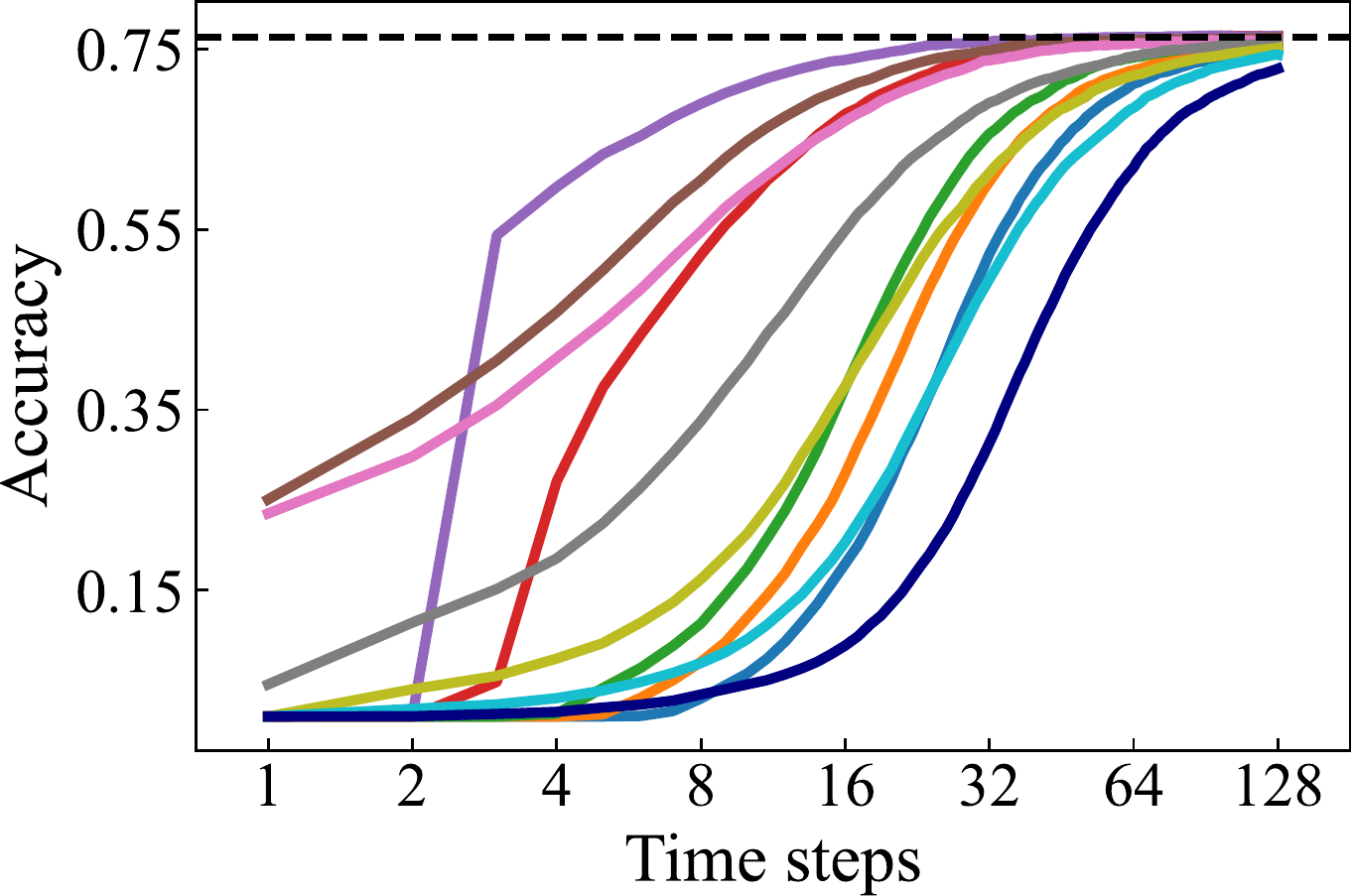}}
 \subfigure[ResNet-20 on CIFAR-100]{\includegraphics[width=0.226\textwidth]{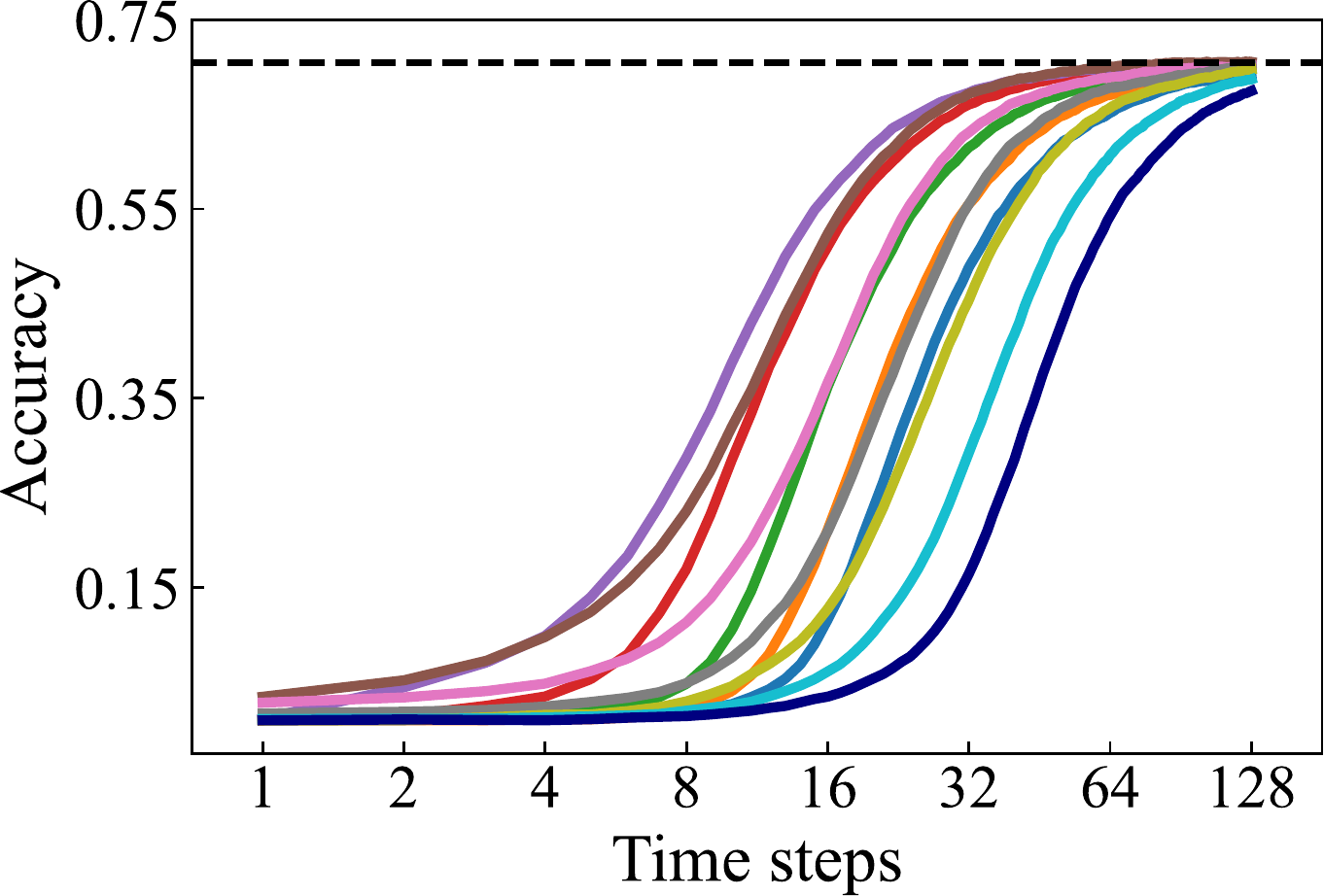}}
 \caption{Comparison of different constant initial membrane potentials with VGG-16/ResNet-20 network structures on CIFAR-10/CIFAR-100 datasets. The dotted line represents the accuracy of source ANN.}
 \label{Figure 3}
\end{figure}

% \begin{figure}[t]
%  \centering
%  \subfigure[VGG-16 on CIFAR-10]{\includegraphics[width=0.226\textwidth]{./figures/vgg10-hot.pdf}}
%  \subfigure[ResNet-20 on CIFAR-10]{\includegraphics[width=0.226\textwidth]{./figures/res10-hot.pdf}}
%  \subfigure[VGG-16 on CIFAR-100]{\includegraphics[width=0.226\textwidth]{./figures/vgg100-hot.pdf}}
%  \subfigure[ResNet-20 on CIFAR-100]{\includegraphics[width=0.226\textwidth]{./figures/res100-hot.pdf}}
%  \caption{Comparison of different constant initial membrane potentials with VGG-16/ResNet-20 network structures on CIFAR-10/CIFAR-100 datasets. 
%  %The dotted line represents the accuracy of source ANN. The change in color in the hotmap indicates the change in the accuracy of the converted SNN.}
%  \label{Figure 3}
% \end{figure}

\begin{table*}[t]
\centering
\scalebox{1}
{
\begin{threeparttable}
\begin{tabular}{@{}lllllllll@{}}
\toprule
Method                                & ANN Acc. & T=8  & T=16  & T=32  & T=64  & T=128 & T=256 & T$\geq$512\\ \midrule[1pt]
\multicolumn{9}{c}{\textbf{VGG-16 \cite{simonyan2014very} on CIFAR-10}}                              \\ \midrule[1pt]
Robust Norm \cite{rueckauer2017conversion} \tnote{1} & 92.82    & -     & 10.11 & 43.03 & 81.52 & 90.80 & 92.75 & 92.75              \\ \midrule 
Spike Norm \cite{sengupta2019going}     & 91.70    & -     & -     & -     & -     & -     & -   & 91.55               \\ \midrule
Hybrid Train \cite{rathi2019enabling}   & 92.81    & -     & -     & -     & -     & 91.13 & -   & 92.48               \\ \midrule
RMP \cite{han2020rmp}                   & 93.63    & -     & -     & 60.30 & 90.35 & 92.41 & 93.04 & 93.63               \\ \midrule
TSC \cite{han2020deep}                  & 93.63    & -     & -     & -     & 92.79 & 93.27 & 93.45 & 93.63              \\ \midrule
Opt. \cite{deng2020optimal}             & 95.72    & -     & -     & 76.24 & 90.64 & 94.11 & 95.33 & 95.73              \\ \midrule
RNL \cite{ding2021optimal}              & 92.82    & -     & 57.90 & 85.40 & 91.15 & 92.51 & 92.95 & 92.95              \\ \midrule
Calibration \cite{li2021free}           & 95.72    & -     & -     & 93.71 & 95.14 & 95.65 & 95.79 & 95.79              \\ \midrule
\textbf{Ours}                           & 94.57    & 90.96 & 93.38 & 94.20 & 94.45 & 94.50 & 94.49 & 94.55          \\ \midrule[1pt]
\multicolumn{9}{c}{\textbf{ResNet-20 \cite{he2016deep} on CIFAR-10}}                                       \\ \midrule[1pt]
Spike-Norm \cite{sengupta2019going}     & 89.10    & -     & -     & -     & -     & -     & -   & 87.46              \\ \midrule
Hybrid Train \cite{rathi2019enabling}   & 93.15    & -     & -     & -     & -     & -     & 92.22   & 92.94            \\ \midrule
RMP \cite{han2020rmp}                   & 91.47    & -     & -     & -     & -     & 87.60 & 89.37   & 91.36         \\ \midrule
TSC \cite{han2020deep}                  & 91.47    & -     & -     & -     & 69.38 & 88.57 & 90.10   & 91.42            \\ \midrule
\textbf{Ours}                           & 92.74    & 66.24 & 87.22 & 91.88 & 92.57 & 92.73 & 92.76   & 92.75       \\ \midrule[1pt]
\multicolumn{9}{c}{\textbf{ResNet-18 \cite{he2016deep} on CIFAR-10}}                                       \\ \midrule[1pt]
Opt. \cite{deng2020optimal} \tnote{2}      & 95.46    & -     & -     & 84.06 & 92.48 & 94.68 & 95.30 & 94.42              \\ \midrule
Calibration \cite{li2021free}\tnote{2}& 95.46    & -     & -     & 94.78 & 95.30 & 95.42 & 95.41    & 95.45          \\ \midrule
\textbf{Ours}                         & 96.04    & 75.44 & 90.43 & 94.82 & 95.92 & 96.08 & 96.06    & 96.06          \\ \bottomrule
\end{tabular}
 \begin{tablenotes}
					\footnotesize
					\item[1] Our implementation of Robust Norm.
				 	\item[2] Instead of utilizing the standard ResNet-18 or ResNet-20, they add two more layers to standard ResNet-18.
				\end{tablenotes}
\end{threeparttable}
}
		\caption{Performance comparison between the proposed method and previous work on CIFAR-10 dataset.}
\end{table*}

\begin{table*}[t]
\centering
\scalebox{1}
{
\begin{threeparttable}
\begin{tabular}{@{}lllllllll@{}}
\toprule
Method                                & ANN Acc. & T=8   & T=16  & T=32  & T=64  & T=128 & T=256 &T$\geq$512 \\ \midrule[1pt]
\multicolumn{9}{c}{\textbf{VGG-16 \cite{simonyan2014very} on CIFAR-100}}                             \\ \midrule[1pt]
Spike-Norm \cite{sengupta2019going}     & 71.22    & -     & -     & -     & -     & -     & -      & 70.77               \\ \midrule
RMP \cite{han2020rmp}                   & 71.22    & -     & -     & -     & -     & 63.76 & 68.34 & 70.93               \\ \midrule
TSC \cite{han2020deep}                  & 71.22    & -     & -     & -     & -     & 69.86 & 70.65 & 70.97               \\ \midrule
Opt. \cite{deng2020optimal}             & 77.89    & -     & -     & 7.64  & 21.84 & 55.04 & 73.54 & 77.71               \\ \midrule
Calibration \cite{li2021free}           & 77.89    & -     & -     & 73.55 & 76.64 & 77.40 & 77.68 & 77.87               \\ \midrule
\textbf{Ours}                           & 76.31    & 60.49 & 70.72 & 74.82 & 75.97 & 76.25 &  76.29 & 76.31               \\ \midrule[1pt]
\multicolumn{9}{c}{\textbf{ResNet-20 \cite{he2016deep} on CIFAR-100}}                                      \\ \midrule[1pt]
Spike-Norm \cite{sengupta2019going}     & 69.72    & -     & -     & -     & -     & -     & -      & 64.09               \\ \midrule
RMP \cite{han2020rmp}                   & 68.72    & -     & -     & 27.64 & 46.91 & 57.69 & 64.06 & 67.82               \\ \midrule
TSC \cite{han2020deep}                   & 68.72    & -     & -     & -     & -     & 58.42 & 65.27 & 68.18               \\ \midrule
\textbf{Ours}                           & 70.43    & 23.09 & 52.34 & 67.18 & 69.96 & 70.51 & 70.59 & 70.53               \\ \midrule[1pt]
\multicolumn{9}{c}{\textbf{ResNet-18 \cite{he2016deep} on CIFAR-100}}                                      \\ \midrule[1pt]
Opt. \cite{deng2020optimal}   \tnote{1}     & 77.16    & -     & -     & 51.27 & 70.12 & 75.81 & 77.22 & 77.19               \\ \midrule
Calibration \cite{li2021free} \tnote{1}     & 77.16    & -     & -     & 76.32 & 77.29 & 77.73 & 77.63 & 77.25               \\ \midrule
\textbf{Ours}                               & 79.36    & 57.70 & 72.85  & 77.86 & 78.98 & 79.20 & 79.26 & 79.28               \\ \bottomrule
\end{tabular}
 \begin{tablenotes}
					\footnotesize
					\item[1] Instead of utilizing the standard ResNet-18 or ResNet-20, they add two more layers to standard ResNet-18.
				\end{tablenotes}
\end{threeparttable}
}
		\caption{Performance comparison between the proposed method and previous work on CIFAR-100 dataset.}
\end{table*}

\begin{table*}[t]
\centering
\scalebox{1}
{
\begin{threeparttable}
\begin{tabular}{@{}lllllllll@{}}
\toprule
Method                                & ANN Acc. & T=8   & T=16  & T=32  & T=64  & T=128 & T=256 &T$\geq$512 \\ \midrule[1pt]
\multicolumn{9}{c}{\textbf{VGG-16 on ImageNet}}                                      \\ \midrule[1pt]
Rmp  \cite{han2020rmp}              & 73.49 & -     & -     & -     & -     & -     & 48.32 & 73.09
\\ \midrule
TSC  \cite{han2020deep}             & 73.49 & -     & -     & -     & -     & -     & 69.71 & 73.46
\\ \midrule
Opt. \cite{deng2020optimal}         & 75.36 & -     & -     & 0.114 & 0.118 & 0.122 & 1.81  & 73.88   \\ \midrule
Calibration(advanced) \cite{li2021free} & 75.36 & -     & -     & 63.64 & 70.69 & 73.32 & 74.23 & 75.32   \\ \midrule
\textbf{Ours}                       & 74.85 & 6.25  & 36.02 & 64.70 & 72.47 & 74.24 & 74.62 & 74.69    \\ \bottomrule
\end{tabular}
\end{threeparttable}
}
		\caption{Performance comparison between the proposed method and previous work on ImageNet}
\end{table*}

\noindent
\textbf{Pre-processing.} We randomly crop and resize the images of CIFAR-10 and CIFAR-100 datasets into shape $32\times32$ after padding 4, and then conduct random horizontal flip to avoid over-fitting. Besides, we use Cutout~\cite{devries2017improved} with the recommended parameters. Specifically, the hole and length are 1 and 16 for CIFAR-10, and 1 and 8 for CIFAR-100. The AutoAugment~\cite{cubuk2019autoaugment} policy is also applied for both datasets. Finally, we apply data normalization on all datasets to ensure that the mean value of all input values is 0 and the standard deviation is 1.
For ImageNet datasets, we randomly crop and resize the image into $224\times224$. We also apply CollorJitter and Label Smooth~\cite{szegedy2016inception} during training. Similar to CIFAR datasets, we normalize all input data to ensure that the mean value is 0 and the standard deviation is 1.

\noindent
\textbf{Hyper-Parameters.}
When training ANNs, we use the Stochastic Gradient Descent optimizer~\cite{bottou2012stochastic} with a momentum parameter of 0.9 and a cosine decay scheduler~\cite{loshchilov2016sgdr} to adjust the learning rate. The initial learning rates for CIFAR-10 and CIFAR-100 are 0.1 and 0.02, respectively. Each model is trained for 300 epochs. For ImageNet dataset, the initial learning rates is set to 0.1 and the total epoch is set to 120.
%The L2-regularization coefficient of clipping upper-bound parameters is properly selected to shorten the inference time in converted SNNs.
The L2-regularization coefficient of the weights and biases is set to $5 \times 10^{-4}$ for CIFAR datasets and $1 \times 10^{-4}$ for ImageNet.
The weight decays of the upper bound parameter $\bm{\theta}$ are $1 \times 10^{-3}$ for VGG-16 on CIFAR-10, $5 \times 10^{-4}$ for ResNet-18/20 on CIFAR-10, VGG-16/ ResNet-18/20 on CIFAR-100, and $1 \times 10^{-4}$ for VGG-16 on ImageNet.

\noindent
\textbf{Training details.}
When evaluating our converted SNN, we use constant input of the test images. %We reproduce the RMP model according to the paper~\cite{han2020rmp}, the performance of which is slighter high than the authors report.
In Fig.~\ref{Figure RMP}, we train ResNet-20 networks on the CIFAR-10 dataset for RMP and RNL, respectively. The RMP model is reproduced according to the paper~\cite{han2020rmp}, and the performance is slighter high than the authors' report. The RNL model~\cite{ding2021optimal} is tested with the codes on GitHub provided by the authors.
All experiments are implemented with PyTorch on a NVIDIA Tesla V100 GPU.

\subsection{The effect of membrane potential initialization}
We first evaluate the effectiveness of the proposed membrane potential initialization. We train an ANN and convert it to four SNNs with different initial membrane potentials. Fig.~\ref{Figure 2} illustrates how the accuracy of converted SNN changes with respect to latency. The blue curve denotes zero initialization, namely without initialization. The orange, green, and red curves denote optimal initialization, random initialization from a uniform distribution, and random initialization from a Gaussian distribution, respectively. One can find that the performance of converted SNNs with non-zero initialization (orange, green and red curves) is much better than that of the converted SNN with zero initialization (blue curve), and the converted SNN with optimal initialization achieves the best performance. Moreover, the SNN with zero initialization cannot work if the latency is fewer than 10 time-steps. 
This phenomenon could be explained as follows.
As illustrated in Fig.~\ref{fig1},  without initialization, the neurons in converted SNN take a long time to fire the first spikes, and thus the network is ``inactive" in the first few time-steps.
When the latency is large enough ($>$256 time-steps), we can find that all these methods can get the same accuracy as source ANN (dotted line).

Then we compare different constant initial membrane potentials, ranging from 0 to 1. The results are shown in Fig.~\ref{Figure 3}.
Overall, a larger time-steps will bring more apparent performance improvement, and the performance of all converted SNNs approach the performance of source ANN with enough time-steps. Furthermore,
we can find that the converted SNNs with non-zero initialization are much better than the converted SNN with zero initialization. 
Note that in this experiment, all thresholds of spiking neurons are resized to 1 by scaling the weights and biases. Thus the theoretically optimal initial membrane potential is 0.5.  The converted SNN from VGG-16/ResNet-20 with an initial membrane potential of 0.5 achieves optimal or near-optimal performance on CIFAR-10 and CIFAR-100 datasets. In fact, the derivations of optimal initial membrane potential are based on the %assumptions that the activations of ANNs are uniformly distributed in every small interval or continuous. 
assumption of piecewise uniform distribution.
Thus a small deviation may be expected, as the assumptions cannot be strictly satisfied. To verify it further, we make a concrete analysis of the performance of converted SNN with different initial values. As shown in Figure 5 in the appendix section, the optimal initial value is always between 0.4 and 0.6. The converted SNN with an initial membrane potential of 0.5 achieves optimal or near-optimal performance.

% In this section, we present a commonly happen phenomenon and explain that phenomenon from a new perspective. In common ANN-SNN conversion methodologies, the converted SNNs often suffers from long inference time to achieve loss-less conversions. We notice that we have to wait a small amount of time before receiving the first spike from the last layer neurons when inferring with converted SNNs, and we always have to wait another period of time until receiving stable spikes from those neurons. 

% Trivially, if we applying threshold balancing on converted SNNs and set the threshold of neurons as maximum value of activation value in ANNs, those neurons need to wait at least one time-step before outputting spikes. Therefore, the propagation delay of this layer is at least one time-step. In a common VGG-16 architecture, as there is 16 layers in total, the least propagation delay is 16 time-steps and the situation may deteriorate in deeper networks.

% One direct cause to that problem is that the all neurons is "inactive" in the first few steps. If we apply proper potential initialization, the neurons can be activated by the initial value and output the first spikes earlier. Therefore, we come up with the method of membrane potential initialization. By setting a initial value between zero and threshold of the neuron in converted SNN, we find a considerable decrease on inference time and a remarkably increase of accuracy in low inference time.

\subsection{Comparison with the State-of-the-Art}
We compare our method to other state-of-the-art ANN-to-SNN conversion methods on the CIFAR-10 dataset, and the results are listed in Table 1. Our model can achieve nearly loss-less conversion with a small inference time. For VGG-16, the proposed method reaches an accuracy of 94.20\% using only 32 time-steps, whereas the methods of Robust Norm, RMP, Opt., RNL and Calibration reach 43.03\%, 60.3\%, 76.24\%, 85.4\% and 93.71\% at the end of 32 time-steps. Moreover, the proposed method achieves an accuracy of 90.96\% using unprecedented 8 time-steps, which is 8 time faster than RMP and Opt. that use 64 time-steps. For ResNet-20, it reaches 91.88\% top-1 accuracy with 32 time-steps. Note that the works of Opt. and Calibration add two layers to standard ResNet-18 rather than use the standard ResNet-18 or ResNet-20 structure. For a fair comparison, we add the experiments of converting an SNN from ResNet-18. For the same time-steps, the performance of our method is much better than Opt., and nearly the same as Calibration, which utilizes advanced pipeline to calibrating the error and adds two more layers. Moreover, we can achieve an accuracy of 75.44\% even the time-steps is only 8. These results show that the proposed method outperforms previous work and can implement fast inference.

Next, we test the performance of our method on the CIFAR-100 dataset. Table 2 compares our method with other state-of-the-art methods.  For VGG-16, the proposed method reaches an accuracy of 74.82\% using only 32 time-steps, whereas the methods of  Opt. and Calibration reach 7.64\% and 73.55\% with the same time-steps. Moreover, for ResNet-20 and ResNet-18, our method can reach 52.34\% and 72.85\% top-1 accuracies, respectively, with 16 time-steps. These results demonstrate that our methods achieve state-of-the-art accuracy, using fewer time-steps.

Finally, we test our method on the ImageNet dataset with VGG-16 architecture. Table 3 compares the results with other state-of-the art methods. Our proposed method can achieve 64.70\% top-1 accuracy using only 32 time-steps and achieve 72.47\% top-1 accuracy with only 64 time-steps. All these results demonstrate that our methods is still effective on very large datasets and can reach state-of-the-art accuracy on all datasets.

\begin{figure}[t]
 \centering
 \subfigure[RMP]{\includegraphics[width=0.23\textwidth]{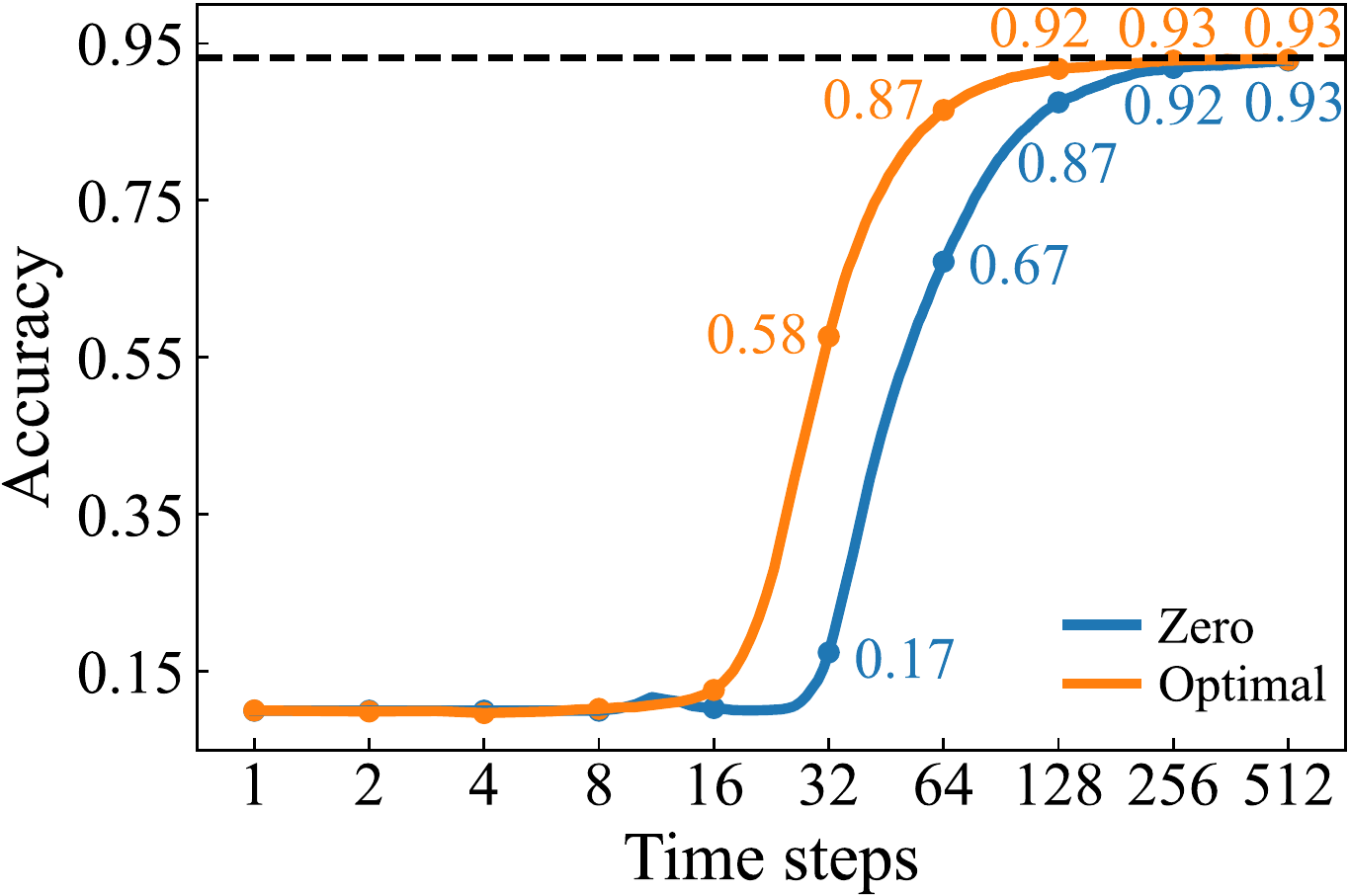}}
 \subfigure[RNL]{\includegraphics[width=0.23\textwidth]{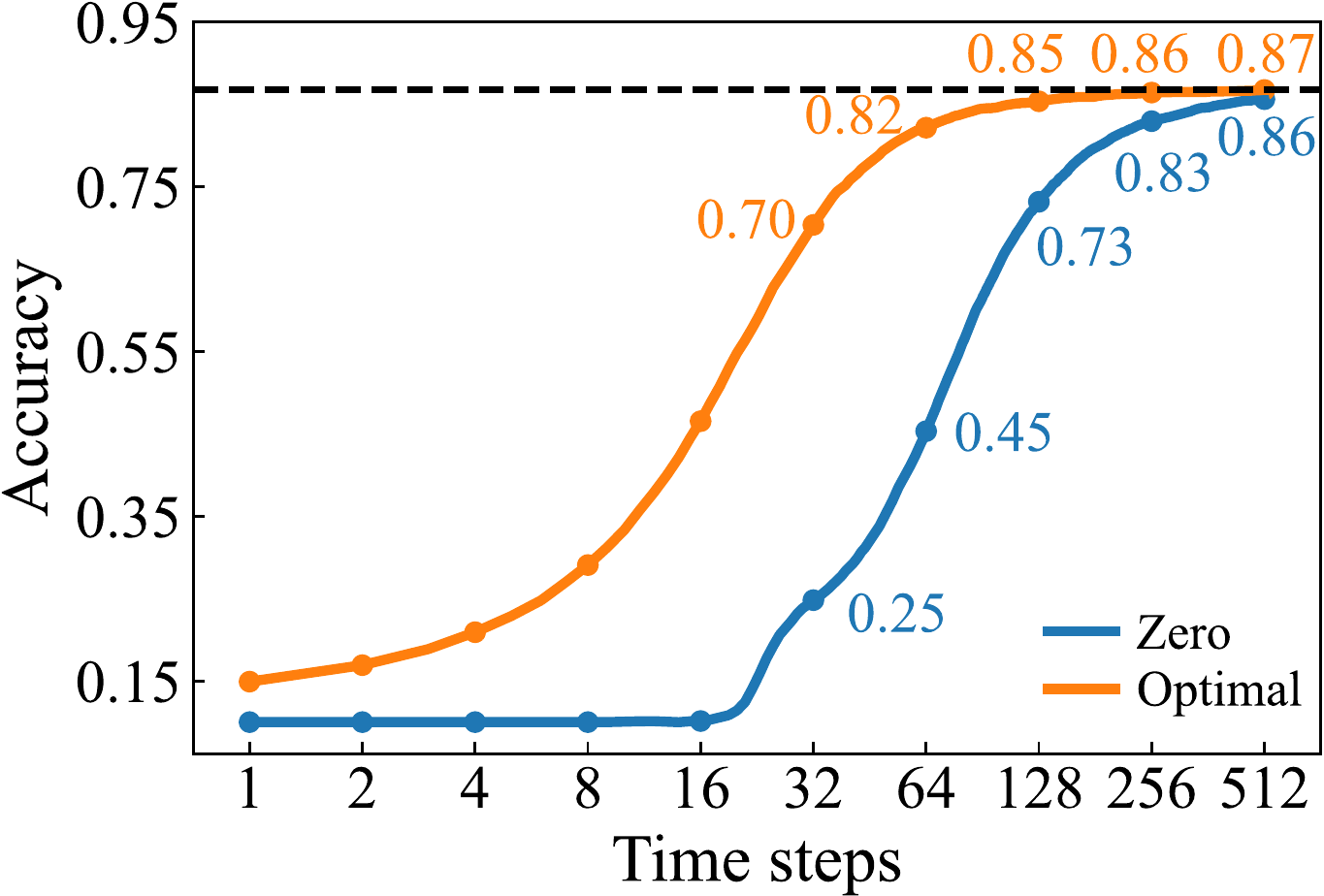}}
 \caption{Comparison of the performance of converted SNNs from ResNet-20 with/without  optimal initial membrane  potential  on the CIFAR-10 dataset.
 }
 \label{Figure RMP}
\end{figure}
\subsection{Apply optimal initial potentials to other models}
Here we test whether the proposed algorithm can be applied to other ANN-to-SNN conversion models. We consider the RMP~\cite{han2020rmp} and RNL~\cite{ding2021optimal} models. We train a ResNet-20 network on the CIFAR-10 dataset for each model and then convert it to two SNNs with/without optimal initial membrane potential.
As illustrated in Fig.~\ref{Figure RMP},
one can find that the performance of converted SNN with optimal initial potential  (orange curve) is much better than SNN without initialization (blue curve). For RMP model, the converted SNN with optimal initialization outperforms the original SNN by 20\% in accuracy (87\% vs 67\%) using 64 time-steps. For RNL model, the converted SNN with optimal initialization outperforms the original SNN by by 37\% in accuracy (82\% vs 45\%) using 64 time-steps. Moreover, the RNL model with optimal initialization outperforms the original SNN by by 45\% in accuracy (70\% vs 25\%) using 32 time-steps.
These results imply that our method is compatible with many ANN-to-SNN conversion methods and can remarkably improve performance when the time-steps is small.

\section{Conclusion}
% In this paper, we theoretically derive the relationship between the activations of ANNs and the dynamics of SNNs, and prove that the initialization of membrane potential can be optimized to implement expected error-free ANN-to-SNN conversion. Besides, we show that the converted SNN with optimal initial potential outperforms state-of-the-art comparing methods on the CIFAR-10 and CIFAR-100 datasets. Moreover, our algorithm is compatible with many ANN-to-SNN conversion methods and can remarkably promote performance in low inference time. 
%We demonstrate that optimal initialization of membrane potentials can not only implement expected error-free ANN-to-SNN conversion, but also reduce the time to the first spike of neurons and thus shortening the inference time.

In this paper, we theoretically derive the relationship between the forwarding process of of an ANN and the dynamics of an SNN. We demonstrate that optimal initialization of membrane potentials can not only implement expected error-free ANN-to-SNN conversion, but also reduce the time to the first spike of neurons and thus shortening the inference time. Besides, we show that the converted SNN with optimal initial potential outperforms state-of-the-art comparing methods on the CIFAR-10, CIFAR-100 and ImageNet datasets. Moreover, our algorithm is compatible with many ANN-to-SNN conversion methods and can remarkably promote performance in low inference time.

\section{Acknowledgments}
This work was supported by the National Natural Science Foundation of China (62176003, 62088102, 61961130392).

\bibliography{aaai22}

\begin{thebibliography}{51}
\providecommand{\natexlab}[1]{#1}

\bibitem[{Bottou(2012)}]{bottou2012stochastic}
Bottou, L. 2012.
\newblock Stochastic gradient descent tricks.
\newblock In \emph{Neural networks: Tricks of the trade}, 421--436. Springer.

\bibitem[{Cao, Chen, and Khosla(2015)}]{cao2015spiking}
Cao, Y.; Chen, Y.; and Khosla, D. 2015.
\newblock Spiking deep convolutional neural networks for energy-efficient
  object recognition.
\newblock \emph{International Journal of Computer Vision}, 113(1): 54--66.

\bibitem[{Chen et~al.(2021)Chen, Yu, Fang, Huang, and Tian}]{ijcai2021-236}
Chen, Y.; Yu, Z.; Fang, W.; Huang, T.; and Tian, Y. 2021.
\newblock Pruning of Deep Spiking Neural Networks through Gradient Rewiring.
\newblock In \emph{International Joint Conference on Artificial Intelligence},
  1713--1721.

\bibitem[{Cubuk et~al.(2019)Cubuk, Zoph, Mane, Vasudevan, and
  Le}]{cubuk2019autoaugment}
Cubuk, E.~D.; Zoph, B.; Mane, D.; Vasudevan, V.; and Le, Q.~V. 2019.
\newblock Autoaugment: Learning augmentation strategies from data.
\newblock In \emph{IEEE Conference on Computer Vision and Pattern Recognition},
  113--123.

\bibitem[{Davies et~al.(2018)Davies, Srinivasa, Lin, Chinya, Cao, Choday,
  Dimou, Joshi, Imam, Jain et~al.}]{davies2018loihi}
Davies, M.; Srinivasa, N.; Lin, T.-H.; Chinya, G.; Cao, Y.; Choday, S.~H.;
  Dimou, G.; Joshi, P.; Imam, N.; Jain, S.; et~al. 2018.
\newblock Loihi: A neuromorphic manycore processor with on-chip learning.
\newblock \emph{IEEE Micro}, 38(1): 82--99.

\bibitem[{Deng et~al.(2020)Deng, Wu, Hu, Liang, Ding, Li, Zhao, Li, and
  Xie}]{deng2020rethinking}
Deng, L.; Wu, Y.; Hu, X.; Liang, L.; Ding, Y.; Li, G.; Zhao, G.; Li, P.; and
  Xie, Y. 2020.
\newblock Rethinking the performance comparison between SNNs and ANNs.
\newblock \emph{Neural Networks}, 121: 294--307.

\bibitem[{Deng and Gu(2021)}]{deng2020optimal}
Deng, S.; and Gu, S. 2021.
\newblock Optimal conversion of conventional artificial neural networks to
  spiking neural networks.
\newblock In \emph{International Conference on Learning Representations}.

\bibitem[{DeVries and Taylor(2017)}]{devries2017improved}
DeVries, T.; and Taylor, G.~W. 2017.
\newblock Improved regularization of convolutional neural networks with cutout.
\newblock \emph{arXiv preprint arXiv:1708.04552}.

\bibitem[{Diehl et~al.(2015)Diehl, Neil, Binas, Cook, Liu, and
  Pfeiffer}]{diehl2015fast}
Diehl, P.~U.; Neil, D.; Binas, J.; Cook, M.; Liu, S.-C.; and Pfeiffer, M. 2015.
\newblock Fast-classifying, high-accuracy spiking deep networks through weight
  and threshold balancing.
\newblock In \emph{International Joint Conference on Neural Networks}, 1--8.

\bibitem[{Ding et~al.(2021)Ding, Yu, Tian, and Huang}]{ding2021optimal}
Ding, J.; Yu, Z.; Tian, Y.; and Huang, T. 2021.
\newblock Optimal ANN-SNN conversion for fast and accurate inference in deep
  spiking neural networks.
\newblock In \emph{International Joint Conference on Artificial Intelligence},
  2328--2336.

\bibitem[{Fang et~al.(2021{\natexlab{a}})Fang, Yu, Chen, Huang, Masquelier, and
  Tian}]{fang2021deep}
Fang, W.; Yu, Z.; Chen, Y.; Huang, T.; Masquelier, T.; and Tian, Y.
  2021{\natexlab{a}}.
\newblock Deep residual learning in spiking neural networks.
\newblock In \emph{Thirty-Fifth Conference on Neural Information Processing
  Systems}.

\bibitem[{Fang et~al.(2021{\natexlab{b}})Fang, Yu, Chen, Masquelier, Huang, and
  Tian}]{fang2020incorporating}
Fang, W.; Yu, Z.; Chen, Y.; Masquelier, T.; Huang, T.; and Tian, Y.
  2021{\natexlab{b}}.
\newblock Incorporating learnable membrane time constant to enhance learning of
  spiking neural networks.
\newblock In \emph{Proceedings of the IEEE/CVF International Conference on
  Computer Vision}, 2661--2671.

\bibitem[{Furber et~al.(2012)Furber, Lester, Plana, Garside, Painkras, Temple,
  and Brown}]{furber2012overview}
Furber, S.~B.; Lester, D.~R.; Plana, L.~A.; Garside, J.~D.; Painkras, E.;
  Temple, S.; and Brown, A.~D. 2012.
\newblock Overview of the spinnaker system architecture.
\newblock \emph{IEEE Transactions on Computers}, 62(12): 2454--2467.

\bibitem[{Gerstner and Kistler(2002)}]{gerstner2002spiking}
Gerstner, W.; and Kistler, W.~M. 2002.
\newblock \emph{Spiking Neuron Models: Single Neurons, Populations,
  Plasticity}.
\newblock Cambridge university press.

\bibitem[{Han and Roy(2020)}]{han2020deep}
Han, B.; and Roy, K. 2020.
\newblock Deep spiking neural network: Energy efficiency through time based
  coding.
\newblock In \emph{European Conference on Computer Vision}, 388--404.

\bibitem[{Han, Srinivasan, and Roy(2020)}]{han2020rmp}
Han, B.; Srinivasan, G.; and Roy, K. 2020.
\newblock {RMP-SNN}: Residual membrane potential neuron for enabling deeper
  high-accuracy and low-latency spiking neural network.
\newblock In \emph{IEEE Conference on Computer Vision and Pattern Recognition},
  13558--13567.

\bibitem[{He et~al.(2016)He, Zhang, Ren, and Sun}]{he2016deep}
He, K.; Zhang, X.; Ren, S.; and Sun, J. 2016.
\newblock Deep residual learning for image recognition.
\newblock In \emph{IEEE Conference on Computer Vision and Pattern Recognition},
  770--778.

\bibitem[{Ho and Chang(2020)}]{ho2020tcl}
Ho, N.-D.; and Chang, I.-J. 2020.
\newblock TCL: an ANN-to-SNN conversion with trainable clipping layers.
\newblock \emph{arXiv preprint arXiv:2008.04509}.

\bibitem[{Hu et~al.(2018)Hu, Tang, Wang, and Pan}]{hu2018spiking}
Hu, Y.; Tang, H.; Wang, Y.; and Pan, G. 2018.
\newblock Spiking deep residual network.
\newblock \emph{arXiv preprint arXiv:1805.01352}.

\bibitem[{Hwang et~al.(2021)Hwang, Chang, Oh, Min, Jang, Park, Yu, Lee, and
  Park}]{hwang2021low}
Hwang, S.; Chang, J.; Oh, M.-H.; Min, K.~K.; Jang, T.; Park, K.; Yu, J.; Lee,
  J.-H.; and Park, B.-G. 2021.
\newblock Low-latency spiking neural networks using pre-charged membrane
  potential and delayed evaluation.
\newblock \emph{Frontiers in Neuroscience}, 15: 135.

\bibitem[{Kheradpisheh and Masquelier(2020)}]{kheradpisheh2020temporal}
Kheradpisheh, S.~R.; and Masquelier, T. 2020.
\newblock Temporal backpropagation for spiking neural networks with one spike
  per neuron.
\newblock \emph{International Journal of Neural Systems}, 30(06): 2050027.

\bibitem[{Kim, Kim, and Kim(2020)}]{kim2020unifying}
Kim, J.; Kim, K.; and Kim, J.-J. 2020.
\newblock Unifying activation- and timing-based learning rules for spiking
  neural networks.
\newblock In \emph{Advances in Neural Information Processing Systems},
  19534--19544.

\bibitem[{Kim et~al.(2020)Kim, Park, Na, and Yoon}]{kim2020spiking}
Kim, S.; Park, S.; Na, B.; and Yoon, S. 2020.
\newblock Spiking-YOLO: Spiking neural network for energy-efficient object
  detection.
\newblock In \emph{AAAI Conference on Artificial Intelligence}, 11270--11277.

\bibitem[{Lee et~al.(2020)Lee, Sarwar, Panda, Srinivasan, and
  Roy}]{lee2020enabling}
Lee, C.; Sarwar, S.~S.; Panda, P.; Srinivasan, G.; and Roy, K. 2020.
\newblock Enabling spike-based backpropagation for training deep neural network
  architectures.
\newblock \emph{Frontiers in Neuroscience}, 14.

\bibitem[{Lee, Delbruck, and Pfeiffer(2016)}]{lee2016training}
Lee, J.~H.; Delbruck, T.; and Pfeiffer, M. 2016.
\newblock Training deep spiking neural networks using backpropagation.
\newblock \emph{Frontiers in Neuroscience}, 10: 508.

\bibitem[{Li et~al.(2021)Li, Deng, Dong, Gong, and Gu}]{li2021free}
Li, Y.; Deng, S.; Dong, X.; Gong, R.; and Gu, S. 2021.
\newblock A free lunch from ANN: Towards efficient, accurate spiking neural
  networks calibration.
\newblock In \emph{International Conference on Machine Learning}, 6316--6325.

\bibitem[{Loshchilov and Hutter(2017)}]{loshchilov2016sgdr}
Loshchilov, I.; and Hutter, F. 2017.
\newblock SGDR: Stochastic gradient descent with warm restarts.
\newblock In \emph{International Conference on Learning Representations}.

\bibitem[{Maass(1997)}]{maas1997networks}
Maass, W. 1997.
\newblock {Networks of spiking neurons: the third generation of neural network
  models}.
\newblock \emph{Neural Networks}, 10(9): 1659--1671.

\bibitem[{Merolla et~al.(2014)Merolla, Arthur, Alvarez-Icaza, Cassidy, Sawada,
  Akopyan, Jackson, Imam, Guo, Nakamura et~al.}]{merolla2014million}
Merolla, P.~A.; Arthur, J.~V.; Alvarez-Icaza, R.; Cassidy, A.~S.; Sawada, J.;
  Akopyan, F.; Jackson, B.~L.; Imam, N.; Guo, C.; Nakamura, Y.; et~al. 2014.
\newblock A million spiking-neuron integrated circuit with a scalable
  communication network and interface.
\newblock \emph{Science}, 345(6197): 668--673.

\bibitem[{Mostafa(2017)}]{mostafa2017supervised}
Mostafa, H. 2017.
\newblock Supervised learning based on temporal coding in spiking neural
  networks.
\newblock \emph{IEEE Transactions on Neural Networks and Learning Systems},
  29(7): 3227--3235.

\bibitem[{Neftci, Mostafa, and Zenke(2019)}]{neftci2019surrogate}
Neftci, E.~O.; Mostafa, H.; and Zenke, F. 2019.
\newblock Surrogate gradient learning in spiking neural networks: Bringing the
  power of gradient-based optimization to spiking neural networks.
\newblock \emph{IEEE Signal Processing Magazine}, 36(6): 51--63.

\bibitem[{Pei et~al.(2019)Pei, Deng, Song, Zhao, Zhang, Wu, Wang, Zou, Wu, He
  et~al.}]{pei2019towards}
Pei, J.; Deng, L.; Song, S.; Zhao, M.; Zhang, Y.; Wu, S.; Wang, G.; Zou, Z.;
  Wu, Z.; He, W.; et~al. 2019.
\newblock Towards artificial general intelligence with hybrid Tianjic chip
  architecture.
\newblock \emph{Nature}, 572(7767): 106--111.

\bibitem[{Qiao et~al.(2015)Qiao, Mostafa, Corradi, Osswald, Stefanini,
  Sumislawska, and Indiveri}]{qiao2015reconfigurable}
Qiao, N.; Mostafa, H.; Corradi, F.; Osswald, M.; Stefanini, F.; Sumislawska,
  D.; and Indiveri, G. 2015.
\newblock A reconfigurable on-line learning spiking neuromorphic processor
  comprising 256 neurons and 128{K} synapses.
\newblock \emph{Frontiers in Neuroscience}, 9: 141.

\bibitem[{Rathi et~al.(2020)Rathi, Srinivasan, Panda, and
  Roy}]{rathi2019enabling}
Rathi, N.; Srinivasan, G.; Panda, P.; and Roy, K. 2020.
\newblock Enabling deep spiking neural networks with hybrid conversion and
  spike timing dependent backpropagation.
\newblock In \emph{International Conference on Learning Representations}.

\bibitem[{Roy, Jaiswal, and Panda(2019)}]{roy2019towards}
Roy, K.; Jaiswal, A.; and Panda, P. 2019.
\newblock {Towards spike-based machine intelligence with neuromorphic
  computing}.
\newblock \emph{Nature}, 575(7784): 607--617.

\bibitem[{Rueckauer et~al.(2016)Rueckauer, Lungu, Hu, and
  Pfeiffer}]{rueckauer2016theory}
Rueckauer, B.; Lungu, I.-A.; Hu, Y.; and Pfeiffer, M. 2016.
\newblock Theory and tools for the conversion of analog to spiking
  convolutional neural networks.
\newblock \emph{arXiv preprint arXiv:1612.04052}.

\bibitem[{Rueckauer et~al.(2017)Rueckauer, Lungu, Hu, Pfeiffer, and
  Liu}]{rueckauer2017conversion}
Rueckauer, B.; Lungu, I.-A.; Hu, Y.; Pfeiffer, M.; and Liu, S.-C. 2017.
\newblock Conversion of continuous-valued deep networks to efficient
  event-driven networks for image classification.
\newblock \emph{Frontiers in Neuroscience}, 11: 682.

\bibitem[{Schemmel et~al.(2010)Schemmel, Br{\"u}derle, Gr{\"u}bl, Hock, Meier,
  and Millner}]{schemmel2010wafer}
Schemmel, J.; Br{\"u}derle, D.; Gr{\"u}bl, A.; Hock, M.; Meier, K.; and
  Millner, S. 2010.
\newblock A wafer-scale neuromorphic hardware system for large-scale neural
  modeling.
\newblock In \emph{IEEE International Symposium on Circuits and Systems},
  1947--1950. IEEE.

\bibitem[{Sengupta et~al.(2019)Sengupta, Ye, Wang, Liu, and
  Roy}]{sengupta2019going}
Sengupta, A.; Ye, Y.; Wang, R.; Liu, C.; and Roy, K. 2019.
\newblock Going deeper in spiking neural networks: {VGG} and residual
  architectures.
\newblock \emph{Frontiers in Neuroscience}, 13: 95.

\bibitem[{Sharmin et~al.(2020)Sharmin, Rathi, Panda, and
  Roy}]{sharmin2020inherent}
Sharmin, S.; Rathi, N.; Panda, P.; and Roy, K. 2020.
\newblock Inherent adversarial robustness of deep spiking neural networks:
  Effects of discrete input encoding and non-linear activations.
\newblock In \emph{European Conference on Computer Vision}, 399--414.

\bibitem[{Shrestha and Orchard(2018)}]{shrestha2018slayer}
Shrestha, S.~B.; and Orchard, G. 2018.
\newblock SLAYER: Spike layer error reassignment in time.
\newblock In \emph{Advances in Neural Information Processing Systems},
  1419--1428.

\bibitem[{Simonyan and Zisserman(2014)}]{simonyan2014very}
Simonyan, K.; and Zisserman, A. 2014.
\newblock Very deep convolutional networks for large-scale image recognition.
\newblock \emph{arXiv preprint arXiv:1409.1556}.

\bibitem[{St{\"o}ckl and Maass(2021)}]{stockl2021optimized}
St{\"o}ckl, C.; and Maass, W. 2021.
\newblock Optimized spiking neurons can classify images with high accuracy
  through temporal coding with two spikes.
\newblock \emph{Nature Machine Intelligence}, 3(3): 230--238.

\bibitem[{Szegedy et~al.(2016)Szegedy, Ioffe, Vanhoucke, and
  Alemi}]{szegedy2016inception}
Szegedy, C.; Ioffe, S.; Vanhoucke, V.; and Alemi, A. 2016.
\newblock Inception-v4, inception-resnet and the impact of residual connections
  on learning.
\newblock \emph{arXiv preprint arXiv:1602.07261}.

\bibitem[{Wu et~al.(2021)Wu, Chua, Zhang, Li, Li, and Tan}]{wu2021tandem}
Wu, J.; Chua, Y.; Zhang, M.; Li, G.; Li, H.; and Tan, K.~C. 2021.
\newblock A tandem learning rule for effective training and rapid inference of
  deep spiking neural networks.
\newblock \emph{IEEE Transactions on Neural Networks and Learning Systems},
  1--15.

\bibitem[{Wu et~al.(2018)Wu, Deng, Li, Zhu, and Shi}]{wu2018STBP}
Wu, Y.; Deng, L.; Li, G.; Zhu, J.; and Shi, L. 2018.
\newblock Spatio-temporal backpropagation for training high-performance spiking
  neural networks.
\newblock \emph{Frontiers in Neuroscience}, 12: 331.

\bibitem[{Zenke et~al.(2021)Zenke, Boht{\'e}, Clopath, Com{\c{s}}a, G{\"o}ltz,
  Maass, Masquelier, Naud, Neftci, Petrovici et~al.}]{zenke2021visualizing}
Zenke, F.; Boht{\'e}, S.~M.; Clopath, C.; Com{\c{s}}a, I.~M.; G{\"o}ltz, J.;
  Maass, W.; Masquelier, T.; Naud, R.; Neftci, E.~O.; Petrovici, M.~A.; et~al.
  2021.
\newblock Visualizing a joint future of neuroscience and neuromorphic
  engineering.
\newblock \emph{Neuron}, 109(4): 571--575.

\bibitem[{Zenke and Vogels(2021)}]{zenke2021remarkable}
Zenke, F.; and Vogels, T.~P. 2021.
\newblock The remarkable robustness of surrogate gradient learning for
  instilling complex function in spiking neural networks.
\newblock \emph{Neural Computation}, 33(4): 899--925.

\bibitem[{Zhang and Li(2020)}]{zhang2020temporal}
Zhang, W.; and Li, P. 2020.
\newblock Temporal spike sequence learning via backpropagation for deep spiking
  neural networks.
\newblock In \emph{Advances in Neural Information Processing Systems},
  12022--12033.

\bibitem[{Zheng et~al.(2021)Zheng, Wu, Deng, Hu, and Li}]{zheng2021going}
Zheng, H.; Wu, Y.; Deng, L.; Hu, Y.; and Li, G. 2021.
\newblock Going deeper With directly-trained larger spiking neural networks.
\newblock In \emph{AAAI Conference on Artificial Intelligence}, 11062--11070.

\bibitem[{Zhou et~al.(2021)Zhou, Li, Chen, Chandrasekaran, and
  Sanyal}]{zhou2021temporal}
Zhou, S.; Li, X.; Chen, Y.; Chandrasekaran, S.~T.; and Sanyal, A. 2021.
\newblock Temporal-coded deep spiking neural network with easy training and
  robust performance.
\newblock In \emph{AAAI Conference on Artificial Intelligence}, 11143--11151.

\end{thebibliography}

% \clearpage
\section{Appendix}
\subsection{Proofs of Theorem 1}
\textbf{theorem 1.}
\textit{The expectation of square conversion error (Eq.~13) reaches the minimum value when the initial value $\bm{v}^{l}(0)$ is ${ \bm{V}^l_{th}}/2$, meanwhile the expectation of conversion error reaches $0$, that is:}
\begin{align}
 % \min_{\bm{v}^{l}(0)} E_{\bm{z}}\left\| f\left( \bm{z} \right) -f^{'}\left( \bm{z} \right) \right\| _{2}^{2}&= \frac{{ V^l_{th}}^2}{12T^2},  \\
%\bm{v}^{l \ast}(0) = 
\arg \min_{\bm{v}^{l}(0)} E_{\bm{z}}\left\| f\left( \bm{z} \right) -f^{'}\left( \bm{z} \right) \right\| _{2}^{2}&= \frac{\bm{V}^l_{th}}{2}, \\
\left.
E_{\bm{z}}\left ( f\left( \bm{z} \right) -f^{'}\left( \bm{z} \right) \right )\right|_{\bm{v}^{l }(0)=\frac{\bm{V}^l_{th}}{2}} &=\bm{0}.
\end{align}

\begin{proof}
The expectation of square conversion error (Eq.~13 in the main text) can be rewritten as:
\begin{align}
&~~~~~E_{\bm{z}}\left\| f\left( \bm{z} \right) -f^{'}\left( \bm{z} \right) \right\| _{2}^{2}  \nonumber \\
&= E_{\bm{z}}\left\|\bm{z} - \frac{V^l_{th}}{T} \left \lfloor \frac{ T \bm{z} +\bm{v}^{l}(0)}{ V^l_{th} } \right \rfloor \right\| _{2}^{2}  \nonumber \\
   &=  \sum_{i=1}^N{ E_{\bm{z}_i} \left(     \bm{z}_i - \frac{V^l_{th}}{T} \left \lfloor \frac{ T \bm{z}_i +\bm{v}^{l}_i(0)}{ V^l_{th} } \right \rfloor    \right)^2   },
   \label{S4}
\end{align}
where $ \bm{z}_i$ and $\bm{v}^{l}_i(0)$ denote the $i$-th element in $ \bm{z}$ and $\bm{v}^{l}(0)$, respectively. $N$ is the number of element in $ \bm{z}$, namely, the number of neurons in layer $l$. In order to minimize $E_{\bm{z}}\left\| f\left( \bm{z} \right) -f^{'}\left( \bm{z} \right) \right\| _{2}^{2}$, we just need to minimize each $E_{\bm{z}_i} \left(     \bm{z}_i - \frac{V^l_{th}}{T} \left \lfloor \frac{ T \bm{z}_i +\bm{v}^{l}_i(0)}{ V^l_{th} } \right \rfloor    \right)^2 $ ($i=1,2,...,N$).
As ${\bm{z}_i}$ is uniformly distributed in every small interval $[m_{t}, m_{t+1}]$ with the probability density function $p_i^t$ ($t=0,1,...,T$), where $m_{0}=0, m_{T+1}=V^l_{th}, m_t=\frac{t V^l_{th}-\bm{v}^{l}_i(0)}{T}$ for $t=1,2,...,T$, we have:

\begin{align}
&~~~~~E_{\bm{z}_i} \left(     \bm{z}_i - \frac{V^l_{th}}{T} \left \lfloor \frac{ T \bm{z}_i +\bm{v}^{l}_i(0)}{ V^l_{th} } \right \rfloor    \right)^2 \nonumber\\
   &= \int_{0}^{(V^l_{th}-\bm{v}^{l}_i(0))/T}  p_i^0  \left(x- \frac{V_{th}^l}{T} \left \lfloor \frac{xT+\bm{v}^{l}_i(0)}{V^l_{th}} \right \rfloor  \right)^2\ dx \nonumber\\
    & + \int_{(V^l_{th}-\bm{v}^{l}_i(0))/T}^{(2V^l_{th}-\bm{v}^{l}_i(0))/T} p_i^1 \left (x - \frac{V_{th}^l}{T} \left \lfloor \frac{xT+\bm{v}^{l}_i(0)}{V^l_{th}} \right \rfloor  \right)^2\ dx \nonumber\\
    & +... \nonumber\\
    & + \int_{((T-1)V^l_{th}-\bm{v}^{l}_i(0))/T}^{(TV^l_{th}-\bm{v}^{l}_i(0))/T} p_i^{T-1} \left(x- \frac{V^l_{th}}{T} \left \lfloor \frac{xT+\bm{v}^{l}_i(0)}{V^l_{th}} \right \rfloor \right)^2\ dx \nonumber\\ 
    & + \int_{(TV^l_{th}-\bm{v}^{l}_i(0))/T}^{V^l_{th}} p_i^{T} \left(x - \frac{V^l_{th}}{T} \left \lfloor \frac{xT+\bm{v}^{l}_i(0)}{V^l_{th}} \right \rfloor \right)^2\ dx \nonumber\\
    &=   \left. \frac{p_i^0 x^3}{3}  \right|_{0}^{(V^l_{th}-\bm{v}^{l}_i(0))/T} + \left. \frac{p_i^1 \left(x-\frac{V^l_{th}}{T}\right)^3}{3}  \right|_{(V^l_{th}-\bm{v}^{l}_i(0))/T}^{(2V^l_{th}-\bm{v}^{l}_i(0))/T} \nonumber \\
    &+... \nonumber \\
    &+\left. \frac{p_i^{T-1} \left(x-\frac{(T-1)V^l_{th}}{T}\right)^3}{3}  \right|_{((T-1)V^l_{th}-\bm{v}^{l}_i(0))/T}^{(TV^l_{th}-\bm{v}^{l}_i(0))/T} \nonumber \\
    &+ \left. \frac{p_i^{T} \left(x-\frac{TV^l_{th}}{T}\right)^3}{3}  \right|_{(TV^l_{th}-\bm{v}^{l}_i(0))/T}^{V^l_{th}} \nonumber \\
    &=\left (\sum_{j=0}^{T-1}p_i^{j} \right) \frac{\left(V^l_{th}- \bm{v}^{l}_i(0)\right)^3+\left(\bm{v}^{l}_i(0)\right)^3}{3T^3}.
\end{align}
The last equality holds as $p_i^0=p_i^T$. One can find that $E_{\bm{z}_i} \left(     \bm{z}_i - \frac{V^l_{th}}{T} \left \lfloor \frac{ T \bm{z}_i +\bm{v}^{l}_i(0)}{ V^l_{th} } \right \rfloor    \right)^2 $ ($i=1,2,...,N$) reaches the  minimal when $\bm{v}^{l}_i(0)=\frac{V^l_{th}}{2}$. Thus we can conclude that:
\begin{align}
    \arg \min_{\bm{v}^{l}(0)} E_{\bm{z}}\left\| f\left( \bm{z} \right) -f^{'}\left( \bm{z} \right) \right\| _{2}^{2}&= \frac{\bm{V}^l_{th}}{2}.
\end{align}
Now we compute the expectation of conversion error,
\begin{align}
    &E_{\bm{z}_i} \left(     \bm{z}_i - \frac{V^l_{th}}{T} \left \lfloor \frac{ T \bm{z}_i +\bm{v}^{l}_i(0)}{ V^l_{th} } \right \rfloor    \right) \nonumber \\
     &= \int_{0}^{(V^l_{th}-\bm{v}^{l}_i(0))/T}  p_i^0  \left(x- \frac{V_{th}^l}{T} \left \lfloor \frac{xT+\bm{v}^{l}_i(0)}{V^l_{th}} \right \rfloor  \right)\ dx \nonumber\\
    & + \int_{(V^l_{th}-\bm{v}^{l}_i(0))/T}^{(2V^l_{th}-\bm{v}^{l}_i(0))/T} p_i^1 \left (x - \frac{V_{th}^l}{T} \left \lfloor \frac{xT+\bm{v}^{l}_i(0)}{V^l_{th}} \right \rfloor  \right)\ dx \nonumber\\
    & +... \nonumber\\
    & + \int_{((T-1)V^l_{th}-\bm{v}^{l}_i(0))/T}^{(TV^l_{th}-\bm{v}^{l}_i(0))/T} p_i^{T-1} \left(x- \frac{V^l_{th}}{T} \left \lfloor \frac{xT+\bm{v}^{l}_i(0)}{V^l_{th}} \right \rfloor \right)\ dx \nonumber\\ 
    & + \int_{(TV^l_{th}-\bm{v}^{l}_i(0))/T}^{V^l_{th}} p_i^{T} \left(x - \frac{V^l_{th}}{T} \left \lfloor \frac{xT+\bm{v}^{l}_i(0)}{V^l_{th}} \right \rfloor \right)\ dx \nonumber\\
    &=\left (\sum_{j=0}^{T-1}p_i^{j} \right) \frac{\left(V^l_{th}- \bm{v}^{l}_i(0)\right)^2-\left(\bm{v}^{l}_i(0)\right)^2}{2T^2}.
\end{align}
If $\bm{v}^{l}_i(0)=\frac{V^l_{th}}{2}$, we have $\left(V^l_{th}- \bm{v}^{l}_i(0)\right)^2-\left(\bm{v}^{l}_i(0)\right)^2=0$, thus we can get that $E_{\bm{z}_i} \left(     \bm{z}_i - \frac{V^l_{th}}{T} \left \lfloor \frac{ T \bm{z}_i +\bm{v}^{l}_i(0)}{ V^l_{th} } \right \rfloor    \right)=0$ and $E_{\bm{z}} \left(     \bm{z}_i - \frac{V^l_{th}}{T} \left \lfloor \frac{ T \bm{z}_i +\bm{v}^{l}_i(0)}{ V^l_{th} } \right \rfloor    \right)=0$  ($i=1,2,...,N$). We can conclude that:

\begin{align}
    \left.
E_{\bm{z}}\left ( f\left( \bm{z} \right) -f^{'}\left( \bm{z} \right) \right )\right|_{\bm{v}^{l }(0)=\frac{\bm{V}^l_{th}}{2}} &=\bm{0}.
\end{align}
\end{proof}

\subsection{The effect of membrane potential initialization}
\begin{figure}[t]
 \centering
 \subfigure[VGG-16 on CIFAR-10]{\includegraphics[width=0.22\textwidth]{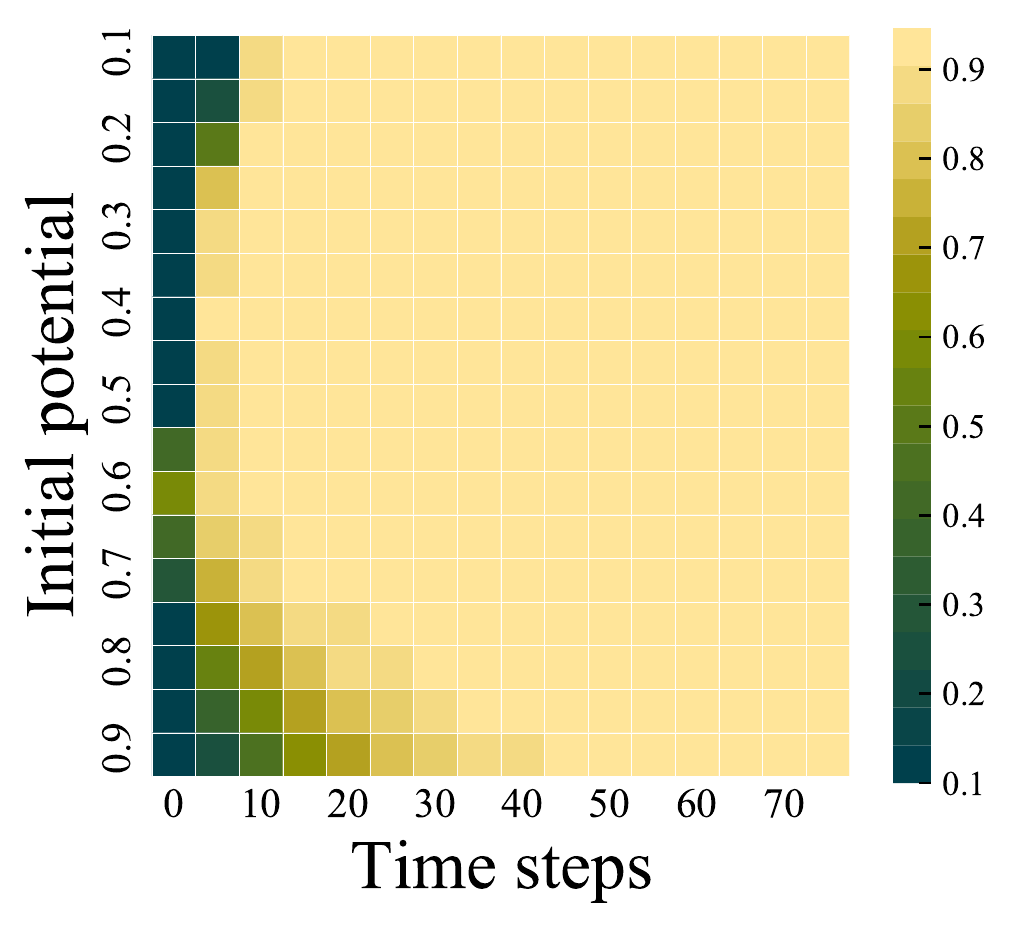}}
 \subfigure[ResNet-20 on CIFAR-10]{\includegraphics[width=0.22\textwidth]{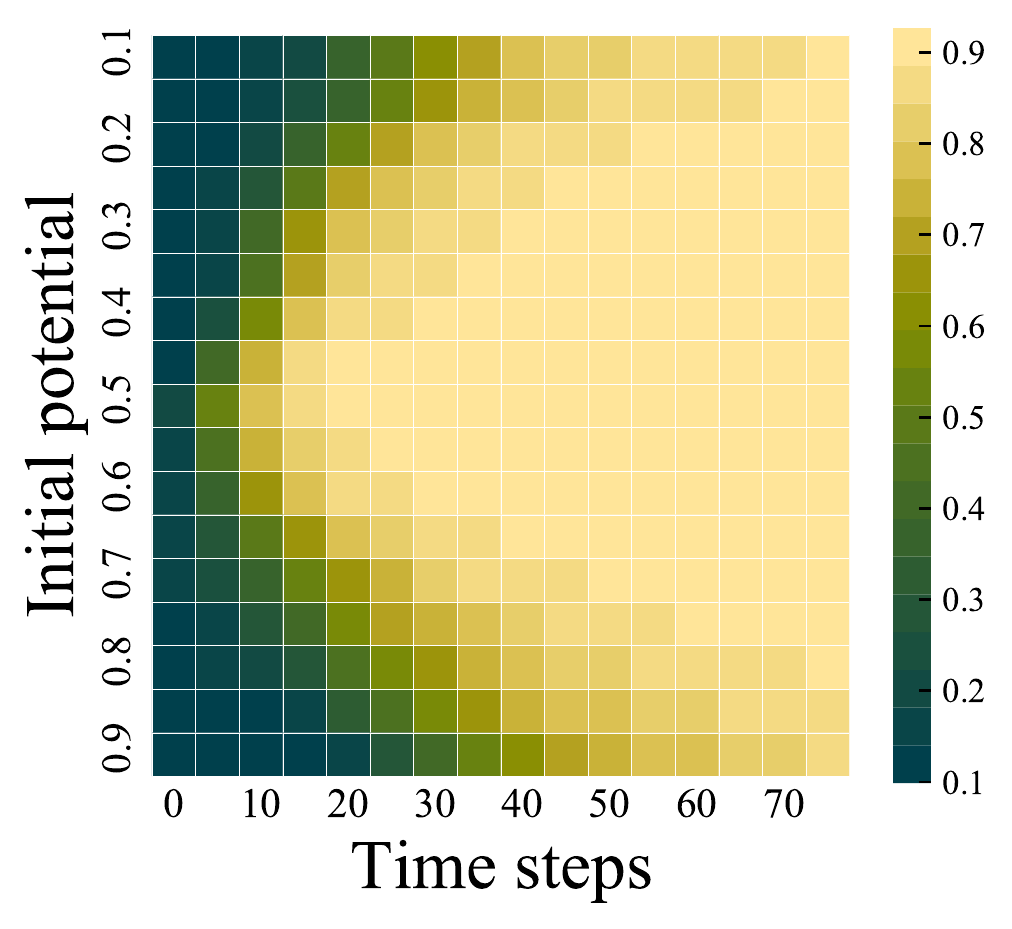}}
 \subfigure[VGG-16 on CIFAR-100]{\includegraphics[width=0.22\textwidth]{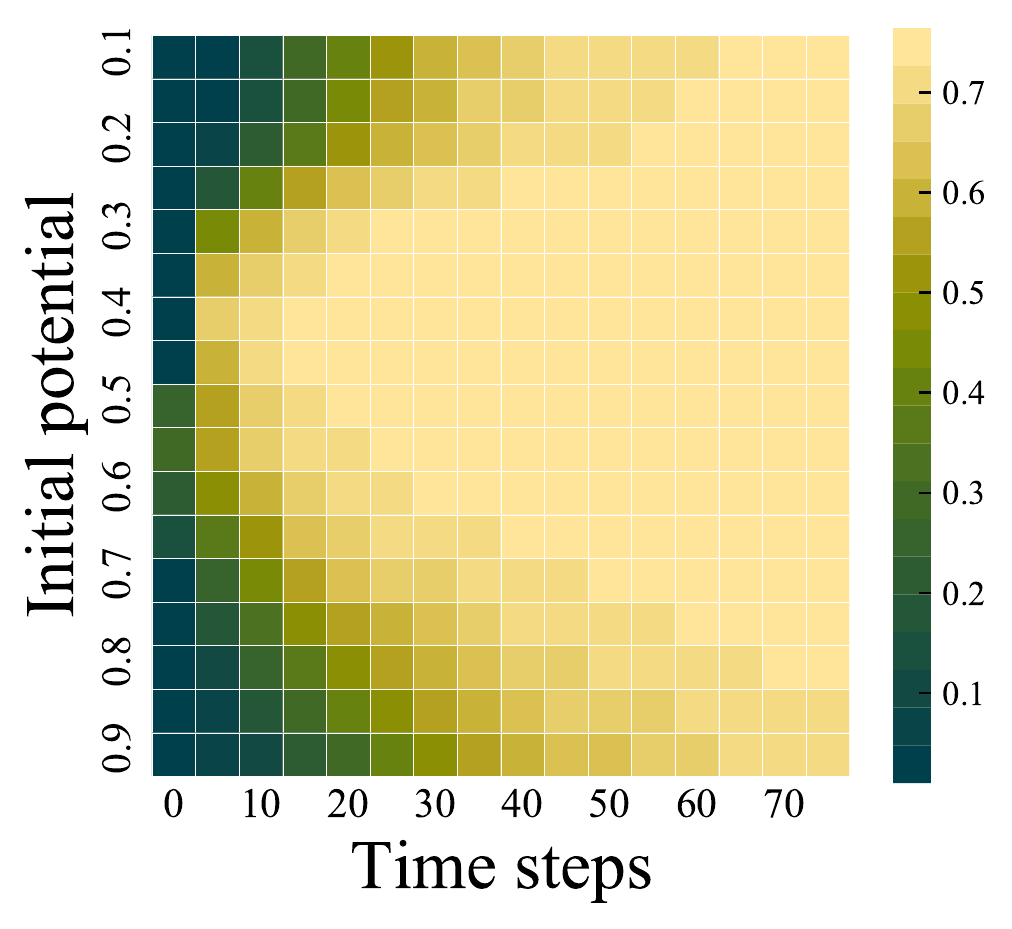}}
 \subfigure[ResNet-20 on CIFAR-100]{\includegraphics[width=0.22\textwidth]{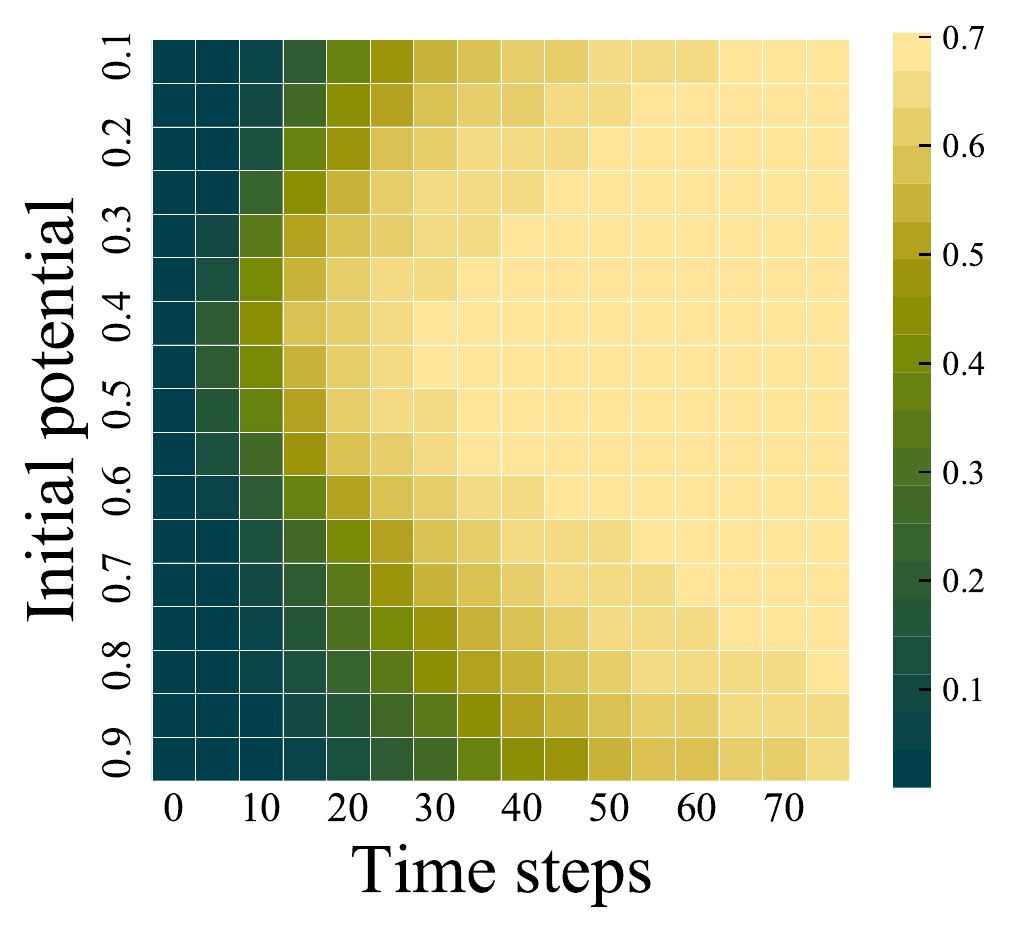}}
 \caption{Performance comparison of different constant initial membrane potentials with VGG-16/ResNet-20 network structures on CIFAR-10/CIFAR-100 datasets. The color represents the accuracy of model.}
 \label{hot maps}
\end{figure}

We make a concrete analysis of the performance of converted SNN with different initial membrane potential, and illustrated the results in Fig.~\ref{hot maps}. Here the times-steps varies from 1 to 75, and the initial potential varies from 0.1 to 0.9. The brighter areas indicate better performance. One can find that the optimal initial potential is always between 0.4 and 0.6, and the converted SNNs with an initial membrane potential of 0.5 achieve optimal or near-optimal performance.

\subsection{Energy Estimation on Neuromorphic Hardware}
% We test the energy consumption on our two models, following the method used in \cite{hu2018spiking}. We use FLOP for ANN and use the synaptic operation SOP for SNN to represent the total numbers of operations for VGG-16 and ResNet-20 to classily one image in CIFAR-10 datasets. We then multiply the number of operations by the power efficiency of FPGAs and neuromorphic hardwraes. For ANN, a Intel Stratix 10 TX operates at a cost of 12.5 pJ per FLOP. For SNN, a neuromorphic chip ROLLS consumes 77fJ per SOP. As the final results listed below, we fixed the inference time of SNN as 32 time steps.
We analyze the energy consumption of our method. Following the analysis method in~\cite{hu2018spiking}, we use FLOP for ANN and the synaptic operation (SOP) for SNN to represent the total numbers of operations to classily one image. We then multiply the number of operations by the power efficiency of FPGAs and neuromorphic hardware, respectively. For ANN, an Intel Stratix 10 TX operates at the cost of 12.5pJ per FLOP, while for SNN, a neuromorphic chip ROLLS consumes 77fJ per SOP~\cite{qiao2015reconfigurable}. Table~\ref{tab:energy} compares the energy consumption of original ANNs (VGG-16 and ResNet-20) and converted SNNs, where the inference time of SNN is set to 32 time-steps. We can find that the proposed method can reach 62 times energy efficiency than ANN with VGG-16 structure and 37 times energy efficiency than ANN with ResNet-20 structure.

\begin{table}[ht]
\centering
\renewcommand\arraystretch{1.2}
\begin{tabular}{lll}
\toprule
                & VGG-16       & ResNet-20       \\ \midrule
ANN OP (MFLOP)  & 332.973      & 41.219          \\ 
SNN OP (MSOP)   & 869.412      & 179.060         \\ 
ANN Power (mJ)  & 4.162        & 0.515           \\ 
SNN Power (mJ)  & 0.067        & 0.0138          \\ 
A/S Power Ratio & 62           & 37              \\ 
\bottomrule
\end{tabular}
\caption{Comparison of power consumption}
\label{tab:energy}
\end{table}

\end{document}